\pdfoutput = 1
\documentclass[sigconf]{acmart}
\renewcommand\footnotetextcopyrightpermission[1]{} % removes footnote with conference information in first column
\usepackage{pgfgantt}
\usepackage{booktabs}
\usepackage{tabulary} %for text tables
\usepackage{graphicx,caption}
\usepackage{mathtools}%loads amsmath
\usepackage{bm}
\usepackage{amssymb,amsfonts}
\usepackage{subcaption}
\usepackage{enumitem}
\usepackage{hyperref}

\setcopyright{none}
\acmConference[Technical Report]{}{}{Boulder, CO}
\acmYear{2018}
\copyrightyear{2018}

\acmPrice{00.00}

\newcommand{\famsec}{FaMSeC}
\newcommand{\solve}{$\mathcal{S}$}
\newcommand{\solvestar}{$\mathcal{S}^*$}
\newcommand{\taskclass}{$c_{\mathcal{T}}$}
\newcommand{\task}{$\mathcal{T}$}

\newcommand{\rwd}{$\mathcal{R}$}
\newcommand{\rwdstar}{$\mathcal{R}^*$}
\newcommand{\rwdstari}{$\mathcal{R}^*_i$}

\newcommand{\rwdstarapprox}{$\widetilde{\mathcal{R}}^*$}
\newcommand{\rwdstariapprox}{$\widetilde{\mathcal{R}}^*_i$}

\newcommand{\surrogate}{$\mathcal{M}^*(\mathcal{T})$}
\newcommand{\xQ}{$x_Q$} %solver quality
\newcommand{\xO}{$x_O$} %outcome assessment
\newcommand{\xP}{$x_P$} %past performance
\newcommand{\xI}{$x_I$}
\newcommand{\xM}{$x_M$}
\newcommand{\xSC}{$x_{SC}$=\{\xI,\xM,\xQ,\xO,\xP \}}

\newcommand{\hell}{$H^2$}
\def\-{\raisebox{.75pt}{-}} %short negative sign
\begin{document}

\title{Factorized Machine Self-Confidence for Decision-Making Agents}%Internal Decision Processes}%%Decision Making Under Uncertainty}%% 
\author{Brett W Israelsen, Nisar R Ahmed, Eric Frew, Dale Lawrence, Brian Argrow} %%nra: I think we should include Eric, Dale and Brian on this as co-authors, since the whole self-confidence block diagram idea came out of deeper discussions with them
    \orcid{0000-0003-1602-1685}
    \email{brett.israelsen@gmail.com}
%\author{Author 2}
    \affiliation{%
        \institution{University of Colorado Boulder}
        \city{Boulder}
        \state{CO}
        \country{USA}
    }
\begin{abstract}
    Algorithmic assurances from advanced autonomous systems assist human users in understanding, trusting, and using such systems appropriately. Designing these systems with the capacity of assessing their own capabilities is one approach to creating an algorithmic assurance. The idea of `machine self-confidence' is introduced for autonomous systems. Using a factorization based framework for self-confidence assessment, one component of self-confidence, called `solver-quality', is discussed in the context of Markov decision processes for autonomous systems. Markov decision processes underlie much of the theory of reinforcement learning, and are commonly used for planning and decision making under uncertainty in robotics and autonomous systems. A `solver quality' metric is formally defined in the context of decision making algorithms based on Markov decisions processes. A method for assessing solver quality  is then derived, drawing inspiration from empirical hardness models. Finally, numerical experiments for an unmanned autonomous vehicle navigation problem under different solver, parameter, and environment conditions indicate that the self-confidence metric exhibits the desired properties. Discussion of results, and avenues for future investigation are included.
\end{abstract}
\maketitle

\section{Introduction}
% Main goal: Argue for SC, and \xQ. Present \xQ{} and justify decisions. Demonstrate the performance on `realistic simulations'.

Thanks to advances in AI and machine learning, unmanned autonomous physical systems (APS) are poised to tackle complex decision making problems for high-consequence applications, such as wilderness search and rescue, transportation, agriculture, remote science, and space exploration. 
%Unlike low-level automation for assembly lines, cruise control, thermostats, etc., 
APS must be self-sufficient and make self-guided decisions about complex problems delegated by users. Hence, APS that are taskable---able to translate high-level commands into suitable processes for sensing, learning, reasoning, communicating, and acting --must also be cognizant and knowledge-rich--capable of reasoning about the capabilities and limitations of their own processes, anticipating possible failures, and able to recognize when they are operating incorrectly \cite{david2016defense}. %to adapt accordingly. %To ensure long-term robustness and resilience for minimally supervised operations, APS behaviors must be predictable, understandable, and explainable. %to human users/stakeholders. %, who in many cases can also provide collaborative high-level assistance or supervisory directives in difficult situations. 

This work is motivated by the need to develop new computational strategies for assessing when an APS reaches its \emph{competency boundaries}. If computed and communicated correctly, such assessments can provide users with clearer predictions of APS behavior and understanding of actual APS capabilities. This can not only allow APS to take initiatives to stay within its competency boundary in untested situations, but also provide users/stakeholders with assurances that allow them to properly calibrate trust in (and hence make proper use of) intelligent APS \cite{Israelsen2017-ym}. 

These properties are especially important for APS that must rely heavily on non-deterministic algorithms for decision-making under uncertainty, i.e. to efficiently make approximate inferences with imperfect models, learn from limited data, and execute potentially risky actions with limited information. 
%From self-driving cars and unmanned aircraft on Earth to planetary rovers in space \cite{}, and from networked smart devices in the home to smart building systems \cite{}, such APS are now becoming a pervasive part of everyday reality. 
%Crucially, these systems require interaction among several different algorithmic components to support intelligent APS capabilities (for sensing, perception, planning, control, communication, etc.). 
%Algorithms for these capabilities are often studied in isolation, but not together... 
%\nisar{but not much work has been done in terms of holistically looking at APS operating under uncertainty -- expand on this in background...} 
Whereas most relevant and recent work on algorithmic introspection and meta-reasoning to date has focused on outcome-based analyses for  AI/learning agents with narrow well-defined tasks,
%such techniques are typically best suited to APS with relatively narrow well-defined capabilities and few computational resource constraints. Since many current and future APS must operate in open-ended task settings in physical environments with significant limitations (due to constrained platform size, weight, power, etc.), the interpretation of `favorable/unfavorable' outcomes can shift in subtle yet significant ways as a function of APS design and task context. To cope with broader classes of APS, %(i.e. engineered sum of interconnected algorithmic and physical parts), 
holistic process-based techniques for algorithmic competency boundary self-assessment are needed to accommodate broader classes of APS operating in complex, dynamic and uncertain real-world settings -- whose computational models and approximations are expected to break down in less obvious/foreknown ways. %\nisar{meh?}

This paper presents and builds on a recently developed algorithmic framework for computing and evaluating self-assessments in APS that leads to shorthand metrics of \emph{machine self-confidence}. Self-confidence is defined as an APS' perceived ability to achieve assigned goals after accounting for uncertainties in its knowledge of the world, its own state, and its own reasoning and execution abilities \cite{Aitken2016-cv, Aitken2016-fb, Sweet2016-tz}. 
Algorithmic computation of self-confidence is strongly linked to model-based assessments of probabilities pertaining to task outcomes and completion---but crucially goes further to provide insight into how well an APS's processes for decision-making, learning, perception, etc. are matched to intended tasks \cite{Hutchins2015-if}. 
We argue that the short-hand insight provided by self-confidence assessments can serve as a transparent and decomposable/traceable feedback signal to anticipate degraded, nominal, or enhanced APS performance, %adapt autonomous behavior, 
and thereby can be used to calibrate user trust in APS for uncertain task settings. 

The main contributions of this paper include: 1) A formal definition of `solver-quality' which is one of several factors that make up `self-confidence'. Herein, solver-quality is presented as a metric for assessing how competent an MDP solver is for a given task. 2) Solver-quality is then derived borrowing inspiration from empirical hardness models (EHMs \cite{Leyton-Brown2009-yr}. 3) Solver-quality is then evaluated using numerical experiments. The paper is organized as follows: In Section~\ref{sec:background} we further explore motivations and background for self-confidence, including concepts like trust between humans and autonomous systems, and a useful example application. In Section~\ref{sec:self-confidence} Factorized Machine Self-Confidence (\famsec) is introduced and a framework outlined. At the end of Section~\ref{sec:self-confidence} we turn our attention to one of the \famsec{} factors, `Solver Quality', and outline specific challenges and desiderata in the context of the broadly useful family of Markov Decision Process (MDP)-based planners. A learning-based technique for computing solver quality factors in MDP-family planners is then derived in Section~\ref{sec:methodology}. In Section~\ref{sec:results} we present results from numerical experiments for an unmanned autonomous vehicle navigation problem. Finally we present conclusions in Section~\ref{sec:conclusions}.

\section{Background and Related Work} \label{sec:background}
This section reviews several key concepts and related works which set the stage for our proposed computational machine self-confidence framework. To make the concepts discussed throughout the paper concrete and provide an accessible proof-of-concept testbed in later sections, we also describe a motivating APS application example inspired by ongoing research in unmanned robotic systems.  

\subsection{Autonomous Systems and User Trust}
An APS is generally any physical agent comprised of a machine controlled by some form of software-based autonomy. Autonomy defines the ability of the system to perform a complex set of tasks with little/no human supervisory intervention for extended periods of time. This generally means that an APS has at least one or more of the capabilities of an artificially intelligent physical agent, i.e. reasoning, knowledge representation, planning, learning, perception, motion/manipulation, and/or communication \cite{Israelsen2017-ym}. 
Despite many popular myths and misconceptions, an APS always interacts with a human user in some way \cite{Bradshaw2013-ck}. 
That is, the aforementioned capabilities are the means by which an APS achieves some \emph{intended} degree of self-sufficiency and self-directedness for tasks that are \emph{delegated} by a user in order to meet an `intent frame' (desired set of goals, plans, constraints, stipulations, and/or value statements) \cite{Miller2014-av}. `Transparency' in this context thus shifts primary concern away from details of how exactly an APS accomplishes a task, towards knowing whether an autonomous system can/cannot execute the task per the user's intent frame. 
In cases where users must re-examine delegated tasks, the ability to interrogate an APS for details related to how tasks would be executed or why tasks can/cannot be completed become an important follow-on consideration for transparency (i.e. on a need to know `drill-down' basis). 
%%\nisar{hook to `drill down' req: user in high-consequence situation would want to know why/why not...and ask more questions as needed or as time/context permits... }

This view naturally sets up several questions related to user trust in autonomous systems. Trust defines a user's willingness and security in depending on an APS to carry out a delegated set of tasks, having taken into consideration its characteristics and capabilities. 
We focus here on the problem of how an APS can be designed to actively assist users in appropriately calibrating their trust in the APS. As surveyed in \cite{Israelsen2017-ym}, several broad classes of \emph{algorithmic assurances} for APS have been developed, where an assurance is defined as any property or behavior that can serve to increase or decrease a user's trust. 
Good assurances are challenging to develop because they must allow users to gain better insight and understanding of APS behaviors for effectively managing operations, without undermining autonomous operations or burdening users in the process. 
Many assurance strategies, such as value alignment \cite{Dragan2014-gu} (where an APS adapts its behavioral objectives with a user's intent frame via interactive learning) and interpretable reasoning \cite{Ruping2006-xj} (where algorithmic capabilities for planning, learning, reasoning, etc. are made accessible and easy to understand for non-expert users) put the onus on the APS (and designers) to integrate naturally transparent trust-calibrating behaviors into core system functionality. 
Other strategies, such as those based on post hoc explanation for learning and reasoning systems \cite{Lacave2004-gq, Ribeiro2016-uc} and data visualization \cite{Sacha2017-hf}, require users to render their own judgments via processing of information provided by the APS (possibly in response to specific user queries). 
Indeed, while the full range of assurance design strategies for APS have much in common with techniques for ensuring transparency and accountability for more general AI and learning systems, assurances based on self-monitoring offer an especially promising path for APS competency assessment. 

\subsection{Self-Monitoring and Self-Confidence}
State of the art machine learning and statistical AI methods have ushered in major improvements to APS capabilities in recent years. 
Yet, as these methods and capabilities continue to improve and find new high-consequence applications, resulting APS implementations are also becoming more complex, opaque and difficult for users (as well as designers and certifying authorities) to fully comprehend. 
In particular, for sophisticated APS characterized by uncertainty-based reasoning and data-driven learning, it can become extremely difficult to make precise predictions about APS behavior and performance limits in noisy, untested, and `out of scope' task conditions with any degree of certainty. Formal verification and validation tools could be used to tackle these issues at design time, but do not provide assurances that can be readily conveyed to or understood by (non-expert) users at run-time. 
It can thus be argued that the task of assessing APS competency at run-time is in general so complex and burdensome that it should also be delegated to the APS itself. 

This leads to consideration of algorithmic self-monitoring methods, e.g. for introspective reasoning/learning \cite{Huang2017-lk}, fault diagnosis and computational meta-reasoning/meta-learning \cite{grant2018recasting}. 
While promising for a wide variety of applications, these methods depend heavily on task outcome and performance assessments, and often require data intensive evaluations. 
As such, these methods are often best-suited to APS with narrow, well-defined, capabilities and few computational resource constraints. 
However, many current and future APS must operate in open-ended task settings in physical environments with significant computational limitations (due to constrained platform size, weight, power, etc.). 
The interpretation of `favorable vs. unfavorable' task outcomes can also shift in subtle yet significant ways that may not be obvious to non-expert users, i.e. depending on the interactions of designed APS capabilities and task context (all of which may also change drastically over the course of a given operational instance). 

%\nisar{...makes it difficult for non-expert users to trace and drill down into and understand assessments... } 
%\nisar{TODO: briefly talk about other background things related to transparency for introspection, Dragan's critical states, meta-RL, meta-cog architectures, etc.; ... what are the drawbacks/limitations in context of autonomous systems? what moves us towards something different like self-confidence? key idea: these are largely outcome based and driven by achievement of robustness/resilience for task performance, as opposed to process-based and do not provide clear assessment/ability to quantify/qualify/communicate what APS actually can or cannot accomplish...taking meta RL as example: system will keep trying to do its best to learn... for critical state MDP: presumes that solver/models/reward functions used to come up with solution are in fact appropriate and that capabilities/scope given to system is in fact appropriate (solution/implementation of APS is not placed into context)}

%\nisar{something here to get into the idea of APS being different but not totally unrelated to ML/AI (e.g. consequences of failure might be loss of APS or damage/irreparable loss to surroundings, as well as dynamic settings...) --  (licensing argument, i.e. we don't certify humans, but do put them through a process-based licensing approach? basically: a big question/central tension here is: how to monitor/supervise APS within competency limits without undercutting very point of autonomy itself? )}

These limitations motivate consideration of \emph{process-based assessment} techniques that allow APS to more generally self-qualify their capabilities for a particular task by evaluating and reporting their associated degree of `self-trust' or \emph{self-confidence}. 
As evidenced by recent work in neurocomputational modeling of decision-making for visual-motor tasks,
self-confidence reporting in humans generally requires second-order reasoning about uncertainties associated with particular task outcomes, i.e. assessments of `uncertainties in uncertainty' as well as of one's own reasoning processes \cite{Adler2016-oi}.  This resonates with the machine self-confidence concept put forth by \cite{Hutchins2015-if}, who proposed using human expert evaluations of specific APS capabilities to manually encode where and when these may break down in particular tasking situations. 
Several formal definitions and techniques for allowing APS to automatically compute their own machine self-confidence scores in the context of different tasks and capability assessments have been proposed recently.
For instance, Kuter and Miller \cite{Kuter2015-qh} proposed to evaluate \emph{plan stability} for systems that rely on hierarchical task planning algorithms, using formal counter-planning methods to determine threatening contingencies for a given plan and plan repair techniques in order to assess the adaptability of that plan to circumvent those contingencies. 
This relies heavily on fixed knowledge bases and ontologies, and so only supports assessments for well-understood environments, tasks, systems, and contexts. 
These and other approaches are reviewed in \cite{Sweet2016-tz}, as well as in \cite{Israelsen2017-ym} in the context of algorithmic interactions for human-autonomous system trust relationships. %, where self-confidence is identified as an explicit assurance in a human-autonomy trust relationship. 
%According to \cite{} the four views on self-confidence are the \textit{anthropomorphic view}, the \textit{uncertainty view}, the \textit{experiential view}, and the \textit{stability view}. The anthropomorphic view defines self-confidence to be similar to how humans express self-confidence, while the experiential view expresses self-confidence based on past experience. The uncertainty view simply defines self-confidence to be the probability of success or failure, and the stability view defines self-confidence to be the sensitivity of the probability of success to uncertainty. All of these views seem to reflect different parts of a more general concept: understanding an autonomous system's ability to do a specific task. 
For the sake of brevity, we restrict attention to the definition of self-confidence used in this work: \textbf{An agent's perceived ability to achieve assigned goals (within a defined region of autonomous behavior) after accounting for (1) uncertainties in its knowledge of the world, (2) uncertainties of its own state, and (3) uncertainties about its reasoning process and execution abilities.}

%\hlr{...also consider other self-confidence ideas/approaches (e.g. four different approaches considered in InfoTech paper) and algorithms (e.g. Ugur's plan robustness idea, surprise index, perception robustness/statistical residuals test)... one limitation of algs: only limited to specific segments or types of reasoning, rather than looking at interconnected processes holistically for autonomous *systems* }

\subsection{MDP-based Planning and Learning}
The diversity of factors that influence APS self-confidence requires a rich modeling approach for algorithmic assessment. We will therefore establish algorithmic realizations of self-confidence assessments by initially studying APS capabilities that can be defined or modeled via Markov decision processes (MDPs). MDPs are composed of finite states and actions that partially resolve the nondeterminism in the state transitions by deciding from what probability distribution $p(\cdot)$ the next state will be sampled. The co-existence of nondeterministic and stochastic choices in MDPs are expressive enough to account for a range of uncertainties including adversarial environmental factors and inaccuracies in execution. %%, and limitations in prior knowledge (e.g. imperfect knowledge of $p(\cdot)$).  
Since MDPs also have well-established connections to other widely used approaches for autonomous decision-making and learning under uncertainty, such as partially observable MDPs (POMDPs) for decision-making in limited observability environments and reinforcement learning for decision-making with incomplete model information \cite{Kochenderfer2015-uu}, they provide an ideal starting point for an initial analysis of self-confidence that can be generalized in future work. 

More formally, we consider generic MDP formulations of a task \task{} delegated to an APS. In an MDP framing of \task{}, the autonomous agent must find an optimal policy $\pi = u(x)$ for an MDP with dynamical state $x$ and actions $u$, such that the objective function
$U = \mathbb{E} \left[\sum_{k=0}^{\infty} \gamma^i r(x_k,u_k) \right]$ is maximized for all times $k=0,...,\infty$ --  
where $R(x_k,u_k)$ rewards (penalizes) the APS for being in (un)desirable states and taking (un)desirable actions, $\mathbb{E}[\cdot]$ is the expected value over all possible future states, and $\gamma \in (0,1]$ is a (tunable) future discount factor. 
Given any $u_k$, the state $x_k$ updates via a Markov probabilistic transition model $x_{k+1} \sim p(x_{k+1}|x_{k},u_{k})$,  
i.e. $x_{i}$ is fully observed at time $i$ (no sensor noise), while transitions $i\rightarrow k+1$ have random perturbations.
In a fully posed MDP, $\pi$ is the optimal state-action policy, which can be recovered from Bellman's equation via dynamic programming. 
However, in many practical situations, policy approximations $\tilde{\pi}$ may still be required, e.g. to cope with very large state dimensions or structured uncertainties in the state transition distribution \cite{Kochenderfer2015-uu}. 
%\nisar{TODO: also help bridge gap to \famsec{} in next section...\famsec{} not exclusive to MDPs, but it's a sensible place to start...note: approximations of reality needed to set up models of decision processes, and then require even more approximations on top of these to actually implement...so this gives a good consequential focus to develop s/c framework}
    
	\begin{figure}[t]%[htbp]
    	\centering
     	\includegraphics[width=0.35\textwidth]{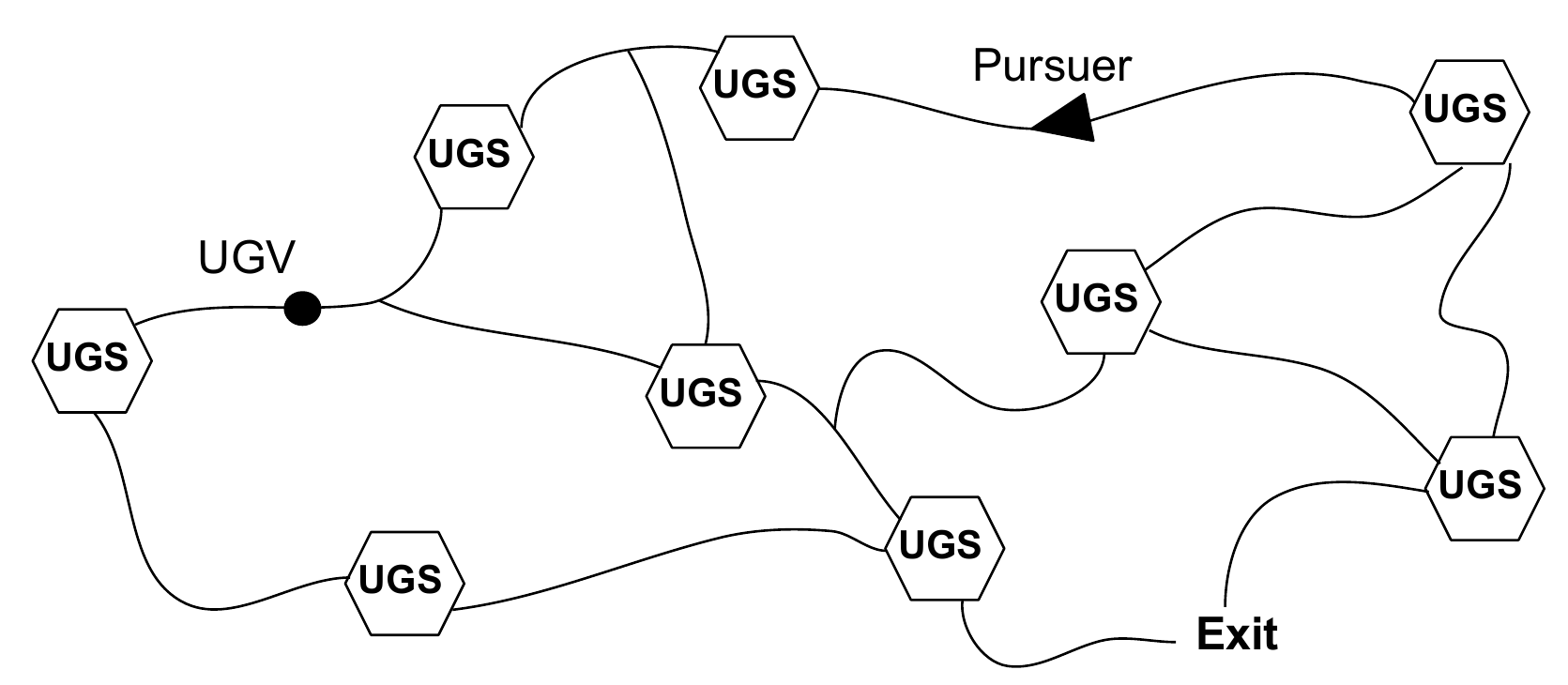}
    	\caption{UGV in road network evading pursuer with information from noisy UGS.} 
        \label{fig:RoadNet}
        \vspace{-0.2cm}
    \end{figure}

\subsubsection{VIP Escort Example Application} \label{sec:vip_escort}
Consider a concrete grounding example problem based on the ``VIP escort'' scenario~\cite{Humphrey2012-lr}, which %can be viewed as a variant of the ``Minotaur's Labyrinth'' and 
serves as a useful proxy for security and surveillance applications with unmanned robotic vehicles (see Fig.~\ref{fig:RoadNet}). An unmanned ground vehicle (UGV) leads a small convoy protecting a VIP through a road network monitored by friendly unattended ground sensors (UGS). The road network also contains a hostile pursuer that the UGV is trying to evade while exiting the network as quickly as possible. The pursuer's location is unknown but can be estimated using intermittent data from the UGS, which only sense portions of the network and can produce false alarms. The UGV's decision space involves selecting a sequence of discrete actions (i.e. go straight, turn left, turn right, go back, stay in place). The UGS data, UGV motion, and pursuer behavior are all stochastic, and the problems of decision making and sensing are strongly coupled: some trajectories through the network allow the UGV to localize the pursuer before heading to the exit but incur a high time penalty); other trajectories afford rapid exit with high pursuer location uncertainty but increase the risk of getting caught by the pursuer, which can take multiple paths. A human supervisor monitors the UGV during operation. The supervisor does not have detailed knowledge of the UGV -- but can interrogate its actions, as well as potentially modify its decision making stance (`aggressively pursue exit' vs. `be very conservative and cautious') %and provide extra information about the pursuer (which is sporadically observed and follows an unknown course). 
in order better cope with the pursuer (which is sporadically observed and follows an unknown course). 

The physical states describing the combined motion of the UGV (whose states are always perfectly observable) and pursuer can be discretized in time and space to produce a Markov process model defined by some initial joint state probability distribution and joint state transition matrix, which depends on the steering actions taken by the UGV. The probability of obtaining `detection' and `no detection' data from each UGS given the true state of the pursuer can be modeled and used to update probabilistic beliefs about the state of the chaser. Finally, a reward function $R(x_k,u_k) = R_k$ can be specified for each time step $k$ to encode user preferences over the combined state of the UGV and pursuer, e.g. $R_k = -1$ for each time step the UGV is not co-located with the pursuer but not yet at the exit, $R_k= -1000$ if the UGV and pursuer are co-located, and $R_k=+1000$ if the UGV reaches the exit without getting caught. Given these elements, the UGV's navigational planning and decision-making problem may be generally formulated as a POMDP. %(more precisely: a mixed observability MDP, if the UGV states are always perfectly known). 
In special the case where the pursuer's state is fully observable at each step $k$ (e.g. due to very reliable and accurate UGS that cover all areas of the road network), the problem reduces to an MDP.

\section{Self-confidence Factorization and Calculation} \label{sec:self-confidence}
    \begin{figure}[tbp]
        \centering
        \includegraphics[width=0.80\linewidth]{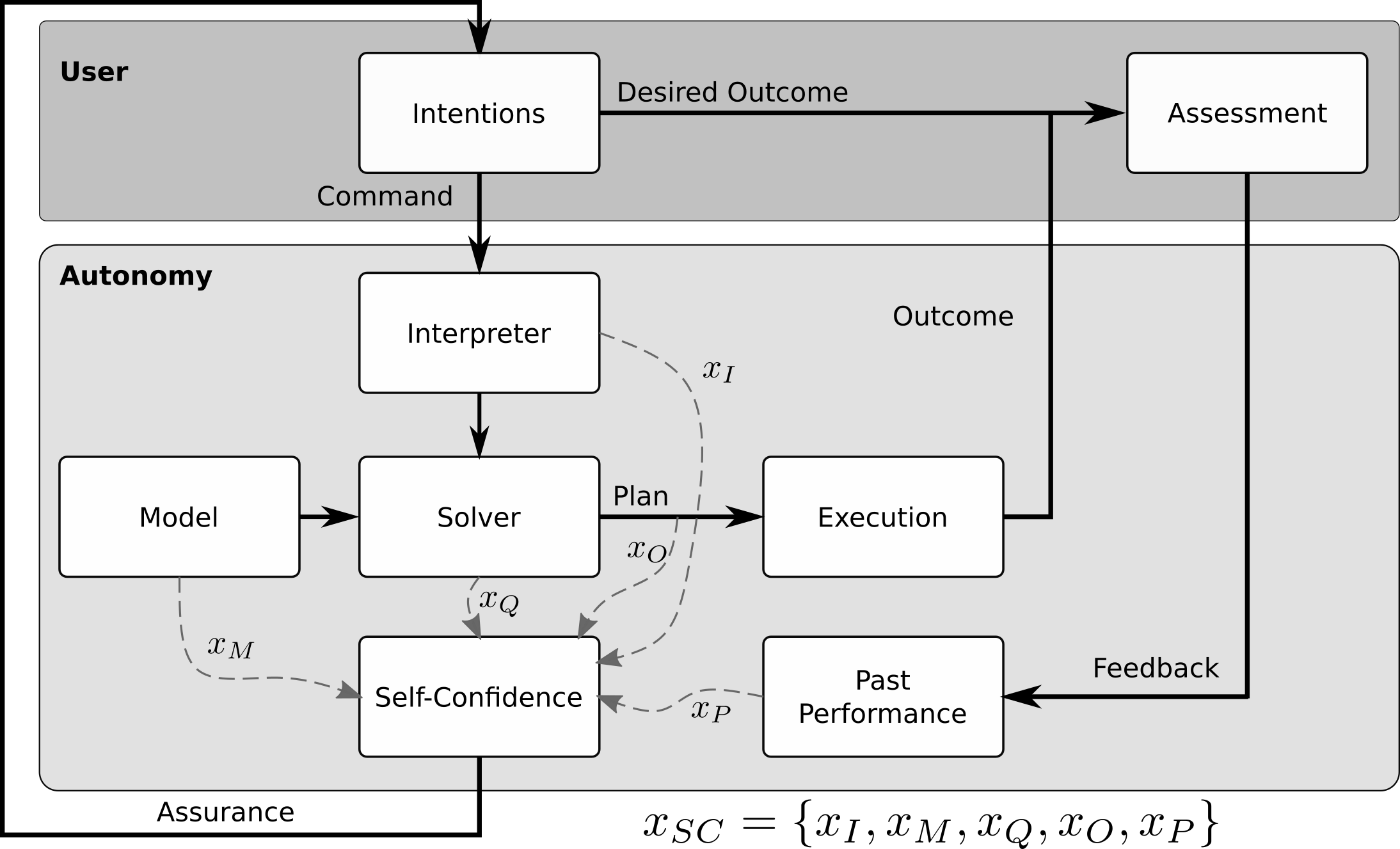}
        \caption{Factorized Machine Self-Confidence (\famsec)}
        \label{fig:famsec}
        \vspace{-0.5cm}
    \end{figure}
    
    This work seeks to develop algorithmic strategies for assessing and communicating machine self-confidence. Of particular interest are model-based techniques that endow an APS with a process-driven scoring of how it arrives at decisions, and what factors influence the quality of its reasoning, in order to quantitatively assess its own competency boundaries. As such, it is important to formally establish both: (i) a set of principles, definitions, and relations that govern the `arithmetic of machine self-confidence' as a function of task, environment, system realization, and context, and (ii) variables, representations and operations for producing meaningful self-confidence assessments. 
    
    We initially address these issues for APS that are primarily defined by capabilities for dynamic decision-making and planning under uncertainty. This approach provides a pathway to developing firm initial mathematical and computational bases for addressing (i) and (ii) via the rich set of analytical and computational features inherent to the MDP model family. %Insights developed along these lines can provide the basis of future work for formulating self-confidence computation strategies, other important planning model families, and APS capabilities that are formally related to decision making under uncertainty, such as dynamic learning and partially observable planning with sensing and perception. 
    After reviewing a computational framework for self-confidence assessment that relies on assessing individual factors involved with solving MDP-based planning and decision-making problems, we consider how one of these factors (related to the quality of a given MDP policy solver) can actually be computed, building on insights derived from calculation and analysis of another factor (related to intrinsic task difficulty) examined in other work. 
    
    \subsection{The \famsec{} Framework }
    The approach presented here adopts and builds on the \emph{Factorized Machine Self-Confidence (\famsec)} framework developed in ref. \cite{Aitken2016-cv, Aitken2016-fb}. The key idea behind \famsec{} is to represent and compute self-confidence as a traceable multi-factor function, which combines shorthand assessments of where and when operations and approximations inherent to model-based autonomous decision-making are expected to break down. As with the self-confidence reporting strategy developed in \cite{Hutchins2015-if}, this captures metrics than an expert designer would use to assess the correctness and quality of an autonomous decision-making system, accounting for variations in task, environment, system implementation, and context. However, unlike \cite{Hutchins2015-if}, \famsec{} allows an APS to automatically generate its own holistic assessments of self-confidence, i.e. without the need for a human designer/expert to specify a priori how self-confident a system ought to be given such variations %%(which can be cumbersome, if not impossible, to fully account for in practical applications). 
    
    Figure \ref{fig:famsec} illustrates \famsec's notional overall self-confidence scoring mechanism. This uses a set of \emph{self-confidence factors} (dashed lines) that are derived from core algorithmic decision-making components (white boxes in the `Autonomy' block). The total self-confidence score can be mapped onto an arbitrary scale, e.g. -1 to +1 for the sake of discussion, where -1 gives a shorthand indication of `complete lack of confidence' (i.e. some aspect of task, environment, or context falls completely outside the system's competency boundaries), and +1 indicates `complete confidence' (i.e. all aspects of task, environment, and mission context are well within system's competency boundaries). As will be shown later, the scales for each factor need not all be the same and can carry slightly different qualitative interpretations, as long as a clear sense of `confidence direction' (i.e. degree of self-trust) can be established for each. 
    Ref. \cite{Aitken2016-cv} considers five general factors that contribute to a `total self-confidence score', which notionally maps the multivariate the combined set of individual factors into an overall confidence report:

\begin{enumerate}
\item \xI---\textit{\textbf{interpretation of user intent and task}}: To what extent were the user's intentions properly understood and translated by the autonomous system into context-appropriate mission specifications and tasks? This factor derives from features and parameters of the `Interpreter' block. For instance, if a natural language interface is used for mission planning, this factor could assess how well user inputs are mapped to reward functions using fixed vocabularies for different mission profiles. 
%This factor can, for instance, capture uncertainty in task objectives or reward functions \cite{abbeel2004apprenticeship, hadfield2016cooperative}. 
%
\item \xM---\textit{\textbf{model and data validity}}: Are the agent's learned and/or assumed models, and associated training data used for decision-making good enough proxies for the real world? This factor assesses how well the set of measurements and events predicted by the autonomous system line up with what it actually should observe in reality. 

%Specifically, this factor uses features and parameters of the `Model' block to assess how well the set of measurements and events predicted by the autonomous system line up with what it actually should observe in reality. 
%For the U2R2 problem, model validity is related to the size of the state space of the model and the level of discretization used in the environment and action set.
%
\item \xQ---\textit{\textbf{solver quality}}: Are the approximations and learning-based adaptations used by the system for solving decision-making problems appropriate for the given mission and model? 
%%This factor uses features and parameters of the `Solver' block to assess the ability of the system to make appropriate decisions based on the information it is given. This factor directly assesses a key aspect of the inherent fitness of the reasoning process: s
Since approximations are almost always needed to solve otherwise intractable decision making problems, this factor examines the appropriateness and reliability of those approximations. 
%For instance, if it is too costly to implement optimal planning, certain problem constraints can often be relaxed to arrive at nearly optimal decisions more cheaply; if such approximations are invalid in critical portions of the problem space, then these can result in poor decisions and hence low confidence in the solution quality. 
This factor also accounts for the impact of learning mechanisms required to make complex decisions under uncertainty, e.g. based on suitability of training data or the learning process to solving the problem at hand. 
%%\nraComm{Ufuk: here is the hook to talk about assessment of learning in MDP families...}
%For the U2R2 problem we will use a priori bounds on the optimality of the resulting policy given known properties of the model and data.
%
\item \xO---\textit{\textbf{expected outcome assessment}}: Do the sets of possible events, rewards, costs, utilities, etc. for a particular decision lead to desirable outcomes? 
Even if the autonomous system perfectly understands and analyzes a task, and can arrive at globally optimal solutions, it may still not be able to always avoid running into undesirable states along the way. % (e.g. due to the existence of catastrophic events with small but non-zero probabilities). 
This factor evaluates the particular decision making strategy implemented by the system to assess the inherent favorability of the full landscape of possible task outcomes.  
%For stochastic policies we use a variation of the upside potential ratio to assess the expected outcome. This measure is further weighted to account for known human biases at extreme outcome probabilities [XYZ]. 
%
\item \xP---\textit{\textbf{past history and experiences}}: What can be gleaned from the system's own experience and other available historical information for past problem instances?  
This factor notionally allows the autonomous system to predict, transfer, and update assessments of self-confidence based on prior experiences, and thus embodies meta-memory and meta-learning for enabling and improving self-assessments. 
%For the U2R2 problem we use a combination of the current learned policy and the rate of change of the policy around the current state to determine how past performance should correlate with current behavior. 
\end{enumerate}

Since the overall self-confidence mapping is heavily dependent on application, context, and desired levels/types of user-autonomy interaction, this work assumes for simplicity that the overall mapping consists of a direct report of some fixed subset of the component factors, i.e. \xSC. 
Furthermore, the five factors considered here are neither exclusive nor exhaustive. For example, the factors developed by \cite{Aitken2016-cv, Aitken2016-fb} are primarily aimed at self-assessment \emph{prior} to the execution of a particular task, whereas it is conceivable that other self-confidence factors could be included to account for in situ and post hoc self-assessments. For simplicity, attention is restricted to the a priori task self-assessment case. %%\nisar{where is traceability mentioned?  notional ability for user to `drill down'?}
%%Incidentally, these characteristics of self-confidence (i.e. self-trust) component factors are not dissimilar from the multidimensional and contextual nature of the component factors that comprise human trust. 

%\nisar{edit...} Self-confidence \xSC{} is a composite assurance \nisar{do we need more jargon?} that stems from some combination of these five factors. 

\subsubsection{VIP Escort Example Revisited}

We will use the VIP Escort scenario to examine two immediate questions: (i) how should the factors be expected to behave under different conditions (independently of how they are actually calculated)?, and (ii) how should any one these factors actually be calculated?  

\begin{figure}[tbp]
    \centering
    \includegraphics[width=0.99\linewidth]{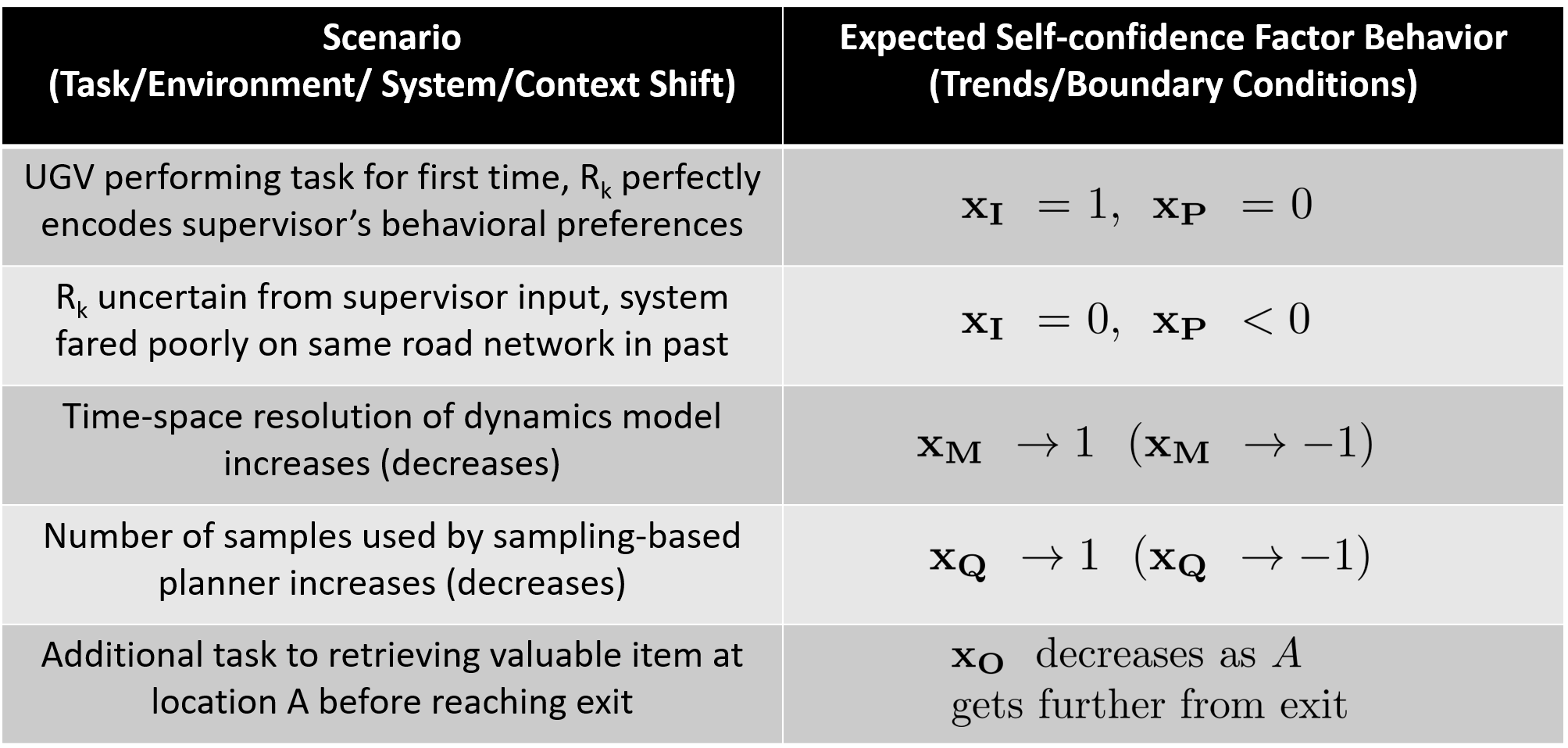}
    \caption{Notional \famsec{} behaviors for VIP Escort problem with a hypothetical sampling-based solver.}
    \label{fig:trendsBCs}
    \vspace{-0.5cm}
\end{figure}

To address (i), we should first consider what kinds of trends, `boundary conditions', and interactions are expected for the various factors if we are given some class of solver for the underlying UGV motion planning problem. For instance, if the problem were modeled and encoded as a discrete-time/discrete-space MDP, then sampling-based Monte Carlo solvers could be used to find an approximately optimal policy $\pi$ \cite{Browne2012-lj}, which would map joint UGV- chaser state information onto specific UGV actions to maximize the UGV's expected cumulative reward. Figure \ref{fig:trendsBCs} shows some expected behaviors for the \famsec{} factors for such a solver, as a function of task, environment, system, and context, assuming again an arbitrary finite range of -1 (total lack of confidence) to +1 (complete confidence) for illustration only. For instance, \xQ{} would be expected to increase/decrease as the number of samples used by the Monte Carlo solver to approximate $\pi$ increased/decreased. Similar trends could also be derived for other non-sampling based solvers.  %\nisar{mention about traceability/drill down basis here?}

With this in mind, an important issue to consider for addressing (ii) is that the factors can depend on each other in complex ways. A logical simplifying assumption for initial algorithm development is thus to consider cases where we can ignore the interactions between factors; this is equivalent to examining each factor along `boundary conditions' where other factors do not change and thus have little/no contribution to the overall self-confidence score. For example, ref. \cite{Aitken2016-cv} developed an approach to compute \xO{} for infinite horizon MDP and POMDP planning, assuming the boundary conditions \xM$=+1$ (perfectly known problem/task model), \xI $= +1$ (perfectly interpreted user task command and reward function $R_k$), \xQ$=+1$ (optimal policy $\pi$ known and available), and \xP$=+1$ (task encountered previously). Under these conditions, overall self-confidence depends only on \xO, which can then be quantified as a measure of the probability distribution $p_{\pi}(R_{\infty})$ of achievable cumulative reward values $R_{\infty} = \sum_{k=0}^{\infty}R_{k}$ under policy $\pi$. Ref. \cite{Aitken2016-cv} considers several measures of $p_{\pi}(R_{\infty})$, including the logistically transformed upper partial moment/lower partial moment (UPM/LPM) score, which quantifies how much probability mass lies to the right vs. left of a minimally acceptable cumulative reward value $R^*_{\infty}$ (e.g. in the basic VIP Escort problem, this corresponds to a user-specified maximum acceptable time to successfully reach the exit).
\begin{figure}[tbp]
    \centering
    \includegraphics[width=0.90\linewidth]{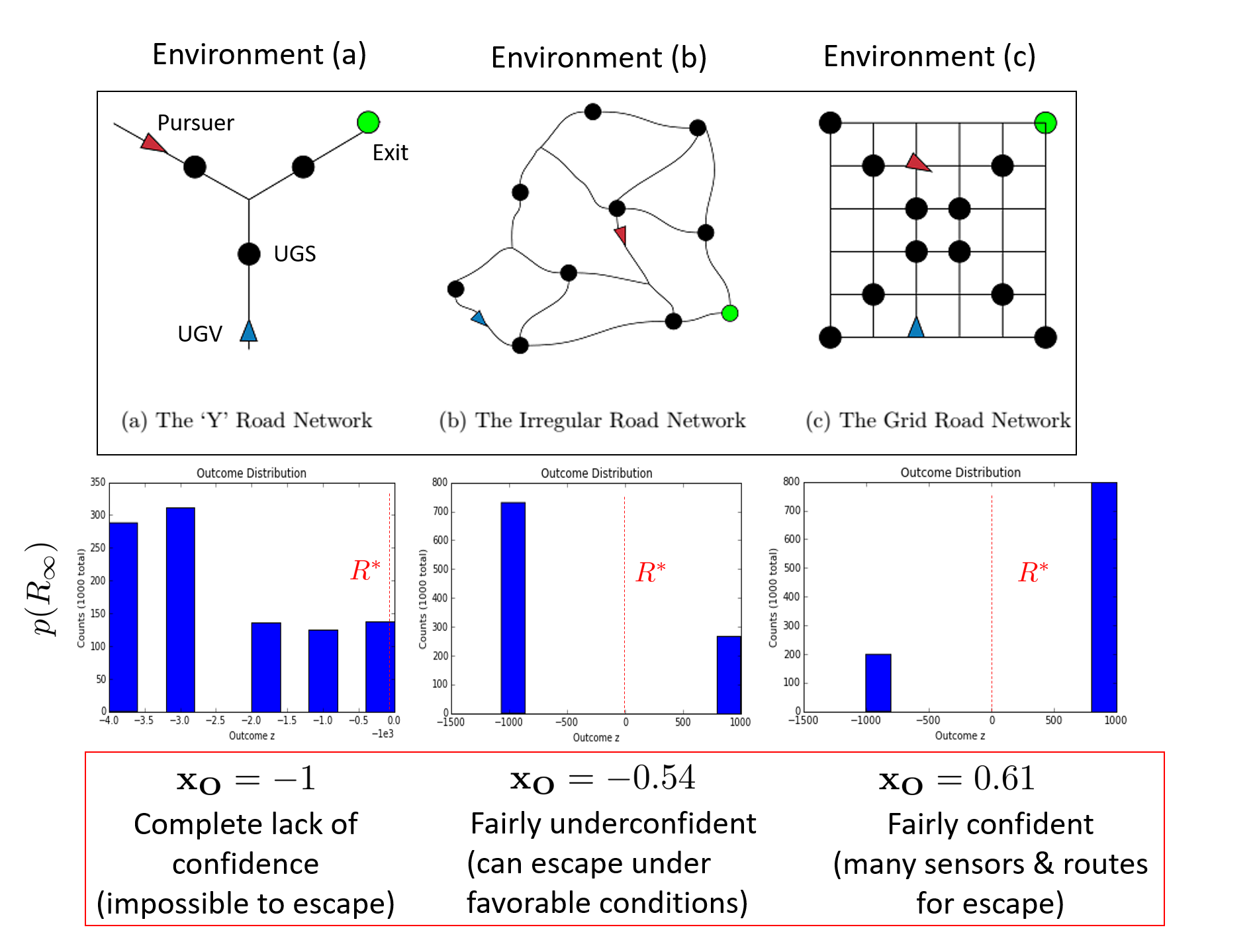}
    \caption{Example \xO{} assessments for VIP Escort problem in various task environments, using UPM/LPM score from \cite{Aitken2016-cv} on empirically sampled $p(R_{\infty})$ pdfs.}
    \label{fig:xOexample}
    \vspace{-0.5cm}
\end{figure}

By indicating how likely favorable outcomes are expected relative to unfavorable outcomes according to a baseline performance measure $R^*_{\infty}$, self-confidence measures like the UPM/LPM score provides information about the consequences of applying policy $\pi$ to a task by interpreting the full shape of the cumulative reward distribution $p_{\pi}(R_{\infty})$, i.e. beyond just the mean value of $R_{\infty}$ (which the optimal $\pi$ maximizes) or the variance/entropy of $p_{\pi}(R_{\infty})$. As illustrated in Fig.~\ref{fig:xOexample} this allows \xO{} to be used as a second-order uncertainty measure for assessing intrinsic task difficulty---and hence indicates a measure of APS competency that can be reported to users to calibrate their trust. %%\nisar{explanations underneath figs not obvious -- mention how user could drill?}

 Since ref. \cite{Aitken2016-cv} does not specify how to compute other factors, nor how to cope with interdependencies between factors that will arise when assumptions such as those above are relaxed, it is natural to consider how these insights extend to computation and analysis of other \famsec{} factors. For instance, what does information related to the assessment of \xO{} tell us about how other factors should be assessed? In particular, since \xO indirectly depends on \xQ, we next consider how to use $p(R_{\infty})$ to also derive a metric for \xQ. Namely, if we consider that an MDP-based APS must in practice often rely on an approximate policy $\tilde{\pi}$ instead of the true optimal policy $\pi$, then a quantitative comparison of $p_{\tilde{\pi}}(R_{\infty})$ to $p_{\pi}(R_{\infty})$ provides a metric for \xQ. The remainder of this paper explores how strategies for assessing \xQ along these lines (under the progressively relaxed assumption of \xM{}=+1, \xP{}=+1, and \xI{}=+1). 

%Ref. \cite{Aitken2016-cv} considers several measures of $p_{\pi}(R_{\infty})$, such as the logistically transformed upper partial moment/lower partial moment (UPM/LPM) score, which quantifies how much probability mass lies to the right vs. left of a minimally acceptable cumulative reward value $R^*_{\infty}$ (e.g. in the basic VIP Escort problem, this corresponds to a user-specified maximum acceptable time to successfully reach the exit). By indicating how likely favorable outcomes are expected relative to unfavorable outcomes according to a baseline performance measure $R^*_{\infty}$, self-confidence measures like the UPM/LPM score provides information about the consequences of applying policy $\pi$ to a task by interpreting the full shape of the cumulative reward distribution $p_{\pi}(R_{\infty})$, i.e. beyond just the mean value of $R_{\infty}$ (which the optimal $\pi$ maximizes) or the variance/entropy of $p_{\pi}(R_{\infty})$. This allows \xO{} to be used a second-order uncertainty measure for assessing intrinsic task difficulty -- and hence indicates a measure of APS competency that can be used to calibrate user trust. \nisar{would including a simple graphical example from Matt's thesis in next section help?}
%%\xO{} was been previously defined in \cite{Aitken2016-cv}, but has not yet been evaluated as an effective assurance as per the guidelines laid out in the work on assurances. Herein \xQ{} is investigated. 

\subsection{Solver Quality (\xQ)} \label{sec:SQ}
    The main aim of \xQ{} is to indicate how a solver \solve{} will perform on a given (possibly un-encountered) task \task{} of a given class \taskclass{} (i.e. all road networks with a UGV, Pursuer, and exit, et cetera as described previously). The need for \xQ{} is not necessarily easy to understand; an analogy helps to clarify:
    
    \emph{Clarifying Example:} One could informally think of \xQ{} as an indication of the \emph{ability} of an athlete. This is opposed to the athlete's assessment of the desirability of the outcome of a game (\xO). While an athlete may be very capable (high \xQ), the score of the game may be such that the athlete knows that it is nearly impossible to catch up and win the game (low \xO). Conversely, an athlete may not be very capable (low \xQ), and due to being na\"{i}ve has an incorrect assessment of the desirability of the outcome (\xO{} cannot be trusted).
    
    The formal desiderata for \xQ{} are:
    
    \begin{enumerate}[label=\textbf{D\arabic*}]
        \item reflect competence of solver \solve{} for task \task{} (where competence is analogous to the `ability' of the athlete in the example)\label{itm:d1}
        \item enable comparison across solver classes \label{itm:d2}
        \item extend to unseen tasks of the same class \taskclass \label{itm:d3}
    \end{enumerate}
    
    For practical application, it is critical to be able to compare the quality of solvers of different classes (i.e. exact vs. approximate) because there are many different ways of solving tasks. Likewise, it is also common for an APS to encounter a similar, but previously unseen, task (i.e. a different road network).
    
    Evaluating the `quality' of something implies some kind of comparison is taking place. In this setting the desired comparison is between a `candidate solver' \solve{} and some reference solver. Ideally, the candidate solver could be compared to the exact solution (whose quality is by definition perfect), but there are three main challenges:

    \begin{enumerate}[label=\textbf{C\arabic*}]
        \item It is unclear how policies/solvers should be compared \label{itm:l1}
        \item Large state spaces make exact solutions infeasible \label{itm:l2}
        \item It is generally impossible to evaluate the exact solution for \emph{all} tasks of a given task class \taskclass{} (linked to \ref{itm:d3}) \label{itm:l3}
    \end{enumerate}

    \subsubsection{Addressing \ref{itm:l1}} \label{sec:compare_policies}
        Solvers of all classes are similar in that they operate on a specified problem in order to produce a policy $\pi$ that is a mapping from states to actions with the aim of maximizing expected reward. A few possibilities for comparing policies include:
    
        \begin{enumerate}
            \item Compare utilities at each state \label{itm:i1}
            \begin{itemize}
                \item Merits: Evaluates whether states are assigned equal utility across solvers. Theoretically state utilities should be independent of the solver. Addresses \ref{itm:d1}
                \item Demerits: Doesn't address \ref{itm:d2}---Doesn't apply when different solvers represent different amounts of the state space, or represent the state space differently.
            \end{itemize} 
            \item Compare `coverage' of the policy (here coverage refers to the proportion of the total state space considered by the solver) \label{itm:i2}
            \begin{itemize}
                \item Merits: Evaluates how `thorough' the policy is; in concert with \ref{itm:i1} could address \ref{itm:d1}
                \item Demerits: Doesn't satisfy \ref{itm:d2}---not all policies have the same coverage, typically by design. Also, high coverage does not imply a `good' solution
            \end{itemize}
            \item Compare the reward distribution of given policies \label{itm:i3}
            \begin{itemize}
                \item Merits: Meets \ref{itm:d1}, also able to satisfy \ref{itm:d2} as reward distributions can be simulated from any policy
                \item Demerits: Expensive to calculate the reward distribution via many simulations
            \end{itemize}
        \end{enumerate}

        Of the possibilities listed above, item \ref{itm:i3} will be used because only it is able to satisfy \ref{itm:d1} and \ref{itm:d2}.

    \subsubsection{Addressing \ref{itm:l2}, and \ref{itm:l3}} \label{sec:practicality}
        In order to address \ref{itm:l2} a `trusted solver' \solvestar{} could be introduced as the reference to which the candidate solver \solve{} can be compared. This solver need not be exact (but could be). Ultimately, \solvestar{} is only required to be a reference of some kind; it may be optimal, or it may be abysmal. In fact given a space of all possible unseen tasks of class \taskclass, \solvestar{} will likely perform very poorly for some of them.

        Still, according to \ref{itm:l3}, it is impractical, or impossible, to find an exact solution for all tasks $\text{task class notation}$. Literature on `Empirical Hardness Models' (EHMs) lends some direction for confronting this challenge. In their work \cite{Leyton-Brown2009-yr,Hutter2009-og} introduced EHMs in order to predict the empirical runtime performance (as opposed to the `Big-O' runtime) of an algorithm on a problem with given features. Specifically, they investigate how the actual runtime of NP-complete problems can be predicted. Applying similar logic in the domain of APS, it should be possible to learn a surrogate model \surrogate{} that predicts the reward distribution \rwdstarapprox{} of a trusted solver \solvestar{} for a given task \task{} of class \taskclass. In this way it is possible to estimate the performance of \solvestar{} on problems to which it has never been applied. This approach also addresses \ref{itm:d3}.
        
    \subsubsection{Summary} The comparison of policies will be done through comparing reward distributions; this approach addresses both \ref{itm:d1} and \ref{itm:d2}, along with \ref{itm:l1}. In order to address \ref{itm:l2}, \ref{itm:l3}, and \ref{itm:d3} a `trusted solver' \solvestar{} will be introduced to serve as a basis by which a `candidate solver' \solve{} can be evaluated. Furthermore, a surrogate model \surrogate{} will be learned to predict \rwdstarapprox{} on un-encountered tasks. In this way, all desiderata, and challenges have been addressed.
        
   \begin{figure}[tb]
        \centering
        \includegraphics[width=0.9\linewidth]{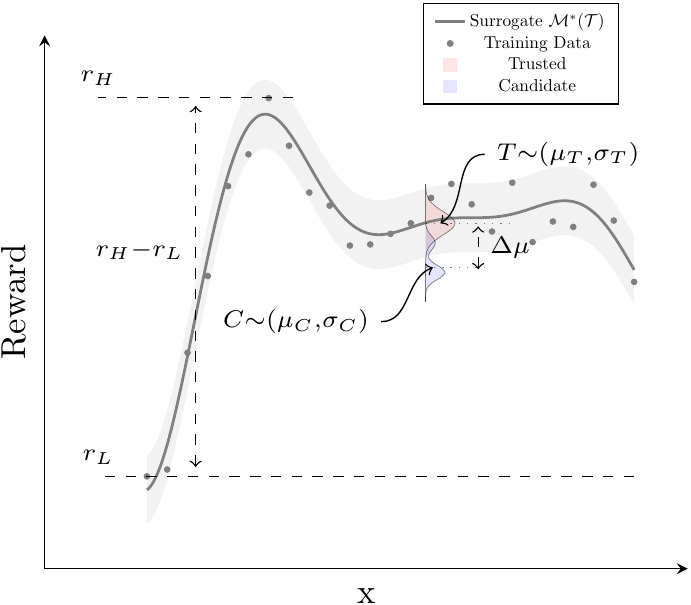}
        \caption{Key values involved in calculating \xQ, where $x$ represents a `parameter of interest' for task \task, or solver \solve.}
        \label{fig:sq_v2}
        \vspace{-0.2cm}
    \end{figure}

\section{Methodology} \label{sec:methodology}
\begin{figure}[tbp]
    \centering
    \begin{subfigure}[c]{0.50\linewidth}
        \centering
        \includegraphics[width=0.75\linewidth]{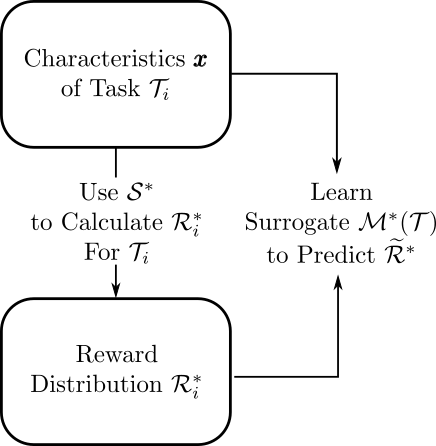}
        \vfill
        \caption{Offline Training}
        \label{fig:sq_train}
    \end{subfigure}%
    \hfill
    \begin{subfigure}[c]{0.50\linewidth}
        \centering
        \includegraphics[width=0.75\linewidth]{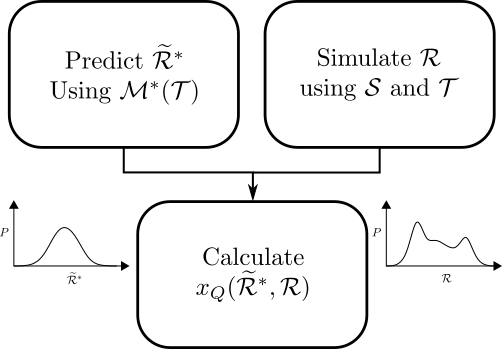}
        \caption{Online Deployment}
        \label{fig:sq_test}
    \end{subfigure} 
    \caption{Depiction of the training phase of the surrogate function \surrogate, and the test, or online deployment, phase where \xQ{} is calculated.}
    \label{fig:sq_test_train}
    \vspace{-0.5cm}
\end{figure}

How, then, can \xQ{} be calculated? Following from discussion in the previous section, a surrogate model \surrogate{} can be learned to predict the reward distribution \rwdstarapprox{} of the trusted reference solver \solvestar{} on task \task{} as shown in Fig.~\ref{fig:sq_train}.

The candidate solver \solve{} must then be evaluated w.r.t. the trusted solver \solvestar{}. This is done by comparing \rwdstarapprox{} (the predicted performance of \solvestar{} on task \task) and \rwd{} (the simulated performance of solver \solve{} on task \task) as illustrated in Fig.~\ref{fig:sq_test}.

Figure \ref{fig:sq_v2} illustrates some of the key quantities involved in calculating \xQ. The basic premise is: \emph{find the difference between the trusted ($T$) and candidate ($C$) solvers while taking into account the overall range of rewards of the trusted solver over many tasks}.

% \begin{figure}[tb]
    % \centering
    % \includegraphics[width=0.95\linewidth]{Figures/SQ_train_test.png}
    % \caption{Depiction of the training phase of the surrogate function \surrogate, and the test, or online deployment, phase where \xQ{} is calculated.}
    % \label{fig:sq_train_test}
% \end{figure}%

\subsection{Learning \surrogate}
The surrogate model \surrogate{} can be any model capable of predicting \rwdstarapprox{} given \task. In the formulation presented above \rwdstariapprox{} only represents the mean and standard deviation for \rwdstari{} (this makes the learning problem less complicated, but is not necessary). Figure \ref{fig:sq_train} depicts how the surrogate model is trained. Learning \surrogate{} would typically be done `offline' and in a supervised manner when more computation power and time are available. Later \surrogate{} can be deployed for use on an APS.

\subsection{Calculating \xQ}
In order to compare two solvers the resultant reward distributions that each of those solvers produce are compared. If two solvers produce an identical reward distribution \emph{for a given task}, then they can be considered equal in their `quality', or considered equally `capable'. Conversely, if the two distributions are very different \emph{for the same task}, then their quality, or capability, is also different.

\subsubsection{Hellinger Metric $\bm{H^2}$} \label{sec:hellinger}
Perhaps the easiest way of calculating the similarity between distributions is to find the `distance' between them, the Hellinger metric (\hell) is such a measure. It is bounded between 0 and 1, where 0 means the distributions are identical. The maximum distance, 1, is achieved when one distribution $P$ assigns zero probability at every point in which another distribution $Q$ assigns probability. \hell{} has different forms based on the type of analytical distributions being compared. For the purposes of calculating \xQ{} from two distributions $P \sim (\mu_1,\sigma_1)$ and $Q\sim(\mu_2,\sigma_2)$ the following form is useful:
\begin{align}
    H^{2}(P,Q) = 1-\sqrt{\frac{2\sigma_P\sigma_Q}{\sigma_P^2+\sigma_Q^2}}\exp{\left(-\frac{1}{4}\frac{(\mu_P-\mu_Q)^2}{\sigma_P^2+\sigma_Q^2}\right)}
\end{align}

Using \hell{} the overlap between $T$ and $C$ can be calculated. However, there are a couple of other considerations that need to be taken into account. 

\subsubsection{Difference in Expected Reward: $\bm{\Delta \mu}$}
\hell{} as a distance measure is always greater than zero, and so information that indicates if a distribution is generally better or worse (i.e. more or less expected reward) is lost. In order to keep this information the sign of the difference between the expected rewards of the two distributions $\text{sgn}(\mu_1-\mu_2)$ can be used.

\subsubsection{Global Scale: $\bm{(r_H-r_L)}$}
The next consideration is that just because distance between two distributions may be great or small, does not mean the same applies from a higher-level, or global, perspective. In an extreme case one might imagine two Normal distributions with means $\mu_1=1$ and $\mu_2=2$, and low variances $\sigma_1^2=\sigma_2^2=1e\-5$. In this case the Hellinger distance between the two would be $1$ since they share practically no overlapping probability. However, if the means of the distributions are nearly equal on the global scale (e.g. rewards from many other training tasks are on the range $[-1e3,1e3]$), then the quantity of \hell{} isn't the critical factor.

\subsubsection{Putting the pieces together}
Using the points discussed above the expression for the quality of candidate solver \solve{} w.r.t. the reference solver \solvestar{} is (again see Fig.~\ref{fig:sq_v2} for intuition):
\begin{align}
    \text{q} &= \text{sgn}(\Delta \mu)f^{\alpha}\sqrt{H^{2}(T,C)} \label{eq:q}
\end{align}

Where $\Delta \mu = \mu_c-\mu_t$, and $f = \Delta \mu/(r_H-r_L)$. The exponent $\alpha$ is a parameter that affects the influence that $f$ has with respect to \hell. In essence, should the relationship of the effects of $f$ and \hell{} be $1:1$? In practice $\alpha=1$ does not yield desirable results. \hell{} should be more influential on $\text{q}$ as $f$ grows smaller, and $f$ should be more influential as it increases. We have found $\alpha=1/2$ gives results that `make sense'; future work could investigate the `best' value for $\alpha$ via user studies.

\subsubsection{Accommodations For Humans}
While \hell{} is on the domain $[0,1]$, the quantity $f$ is $[0,\infty]$. Because of this it is desirable to use a `squashing function' to keep the reported \xQ{} value within some bounded range and avoid arbitrarily large values that can be confusing to humans. The general logistic equation is useful for this.
The numerator is 2 so that when $q=0$ (distributions are identical) \xQ{} will be 1. Dividing the quantity $q$ by 5 so makes it so that \xQ{} `saturates' at around $\text{q}=\pm1$.
\begin{align}
    x_{Q} &= \frac{2}{1+exp(-\text{q}/5)}\label{eq:SQ}
\end{align}

\subsection{Examples}
\begin{figure}[tbp]
    \centering
    \begin{subfigure}[c]{0.65\linewidth}
        \centering
        \includegraphics[width=1.0\linewidth]{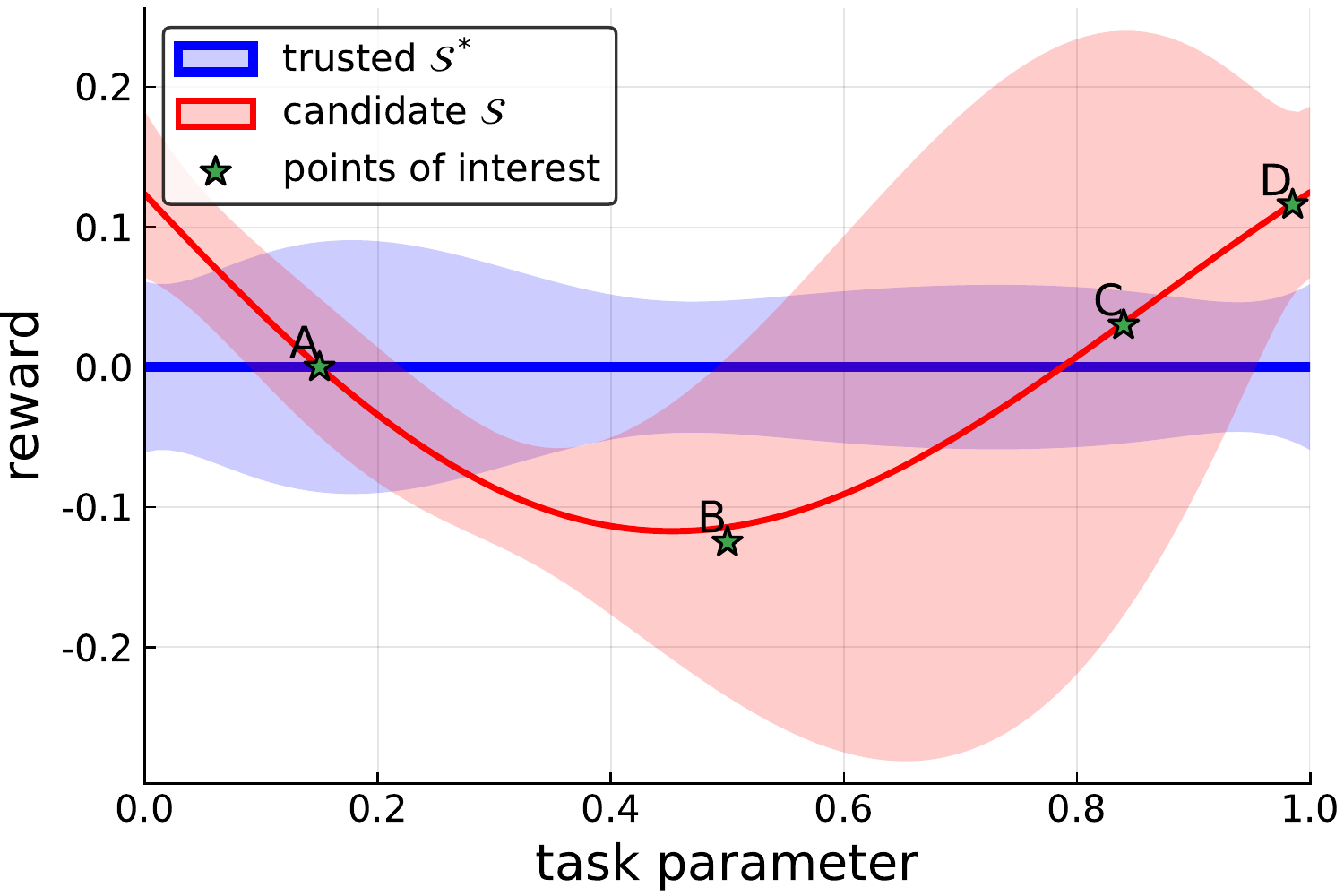}
        \vfill
        % \caption{tst}
        % \label{fig:}
    \end{subfigure}%
    \hfill
    \begin{subfigure}[t]{0.35\linewidth}
        \centering
        \includegraphics[width=1.0\linewidth]{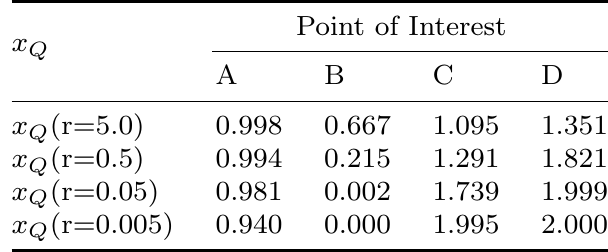}
        % \caption{Candidate solver depth 1}
        % \label{fig:med_roadnet}
    \end{subfigure} 
    \caption{Assessing \xQ{} calculation on reward fxn's: \solvestar{} (blue) and \solve{} (red). Points of interest indicated by a star.}
    \label{fig:sq_thry1}
    \vspace{-0.4cm}
\end{figure}

A toy example is useful in evaluating whether \xQ{} yields desirable results. Figure~\ref{fig:sq_thry1} illustrates a such an example, depicting the expected reward (with uncertainty) for a trusted solver \solvestar{} given a specific, generic, task/solver parameter, as well as that of a `candidate' solver \solve. Different points of interest (indicating specific values of the task parameter) are highlighted by a star. The table on the side shows the values of \xQ{} calculated for different cases.

At B the candidate solver has a lower expected reward than the trusted solver and a higher variance than the trusted solver. Intuitively \xQ{} should be less than one. As shown when $r=5$ (i.e. $r_H-r_L=5$, the global reward range is `large') $x_Q=0.667$ which indicates that the candidate solver is marginally less capable than the trusted solver, and when $r=0.05$ then $x_Q=0.002$ indicates that \solve{} is much less capable than \solvestar.

At C the candidate solver \solve{} has higher expected reward than \solvestar, but a larger variance. Intuitively we would expect \xQ{} of \solve{} to be a little greater than one, and in fact when $r=5$, $x_Q=1.095$. As the global reward range $r$ decreases the difference in capability between \solve{} and \solvestar{} increases with $x_Q=1.995$ at $r=0.005$. These calculations indicate that \xQ{} performs as expected. In Sec.~\ref{sec:results} a more realistic scenario is considered. 

\section{Results} \label{sec:results}
In order to investigate how \xQ{} performs on a more realistic scenario the VIP escort problem introduced in Sec.~\ref{sec:vip_escort} will be used. While the original problem is defined as a POMDP, here we investigate a somewhat simpler version of the problem, and instead use fully observable MDPs. This is reasonable because \xQ{} operates on reward distributions, which are produced by policies on any decision making problem. The benefit of using MDPs is that they are computationally less burdensome than POMDPs, while still being capable of being applied to complex decision problems.

In order to find a policy for the MDP, a Monte-Carlo Tree Search (MCTS) solver will be used \cite{Browne2012-lj}. As the name suggests MCTS involves building a tree from the starting state of the UGV and simulating a specified number of actions/transitions into the future in order to calculate the utility of each state. This process is repeated many times, and the utilities of each state are updated after each iteration. The actions selected by MCTS are based not only on the current utility of the state, but an exploration parameter that helps ensure that the search doesn't simply exploit the greatest known utilities. An MCTS solver is convenient to use during these experiments because the quality of the solver can be easily changed by modifying the parameters.

\begin{figure}[tbp]
    \centering
    \begin{subfigure}[b]{0.50\linewidth}
        \centering
        \includegraphics[width=0.7\linewidth]{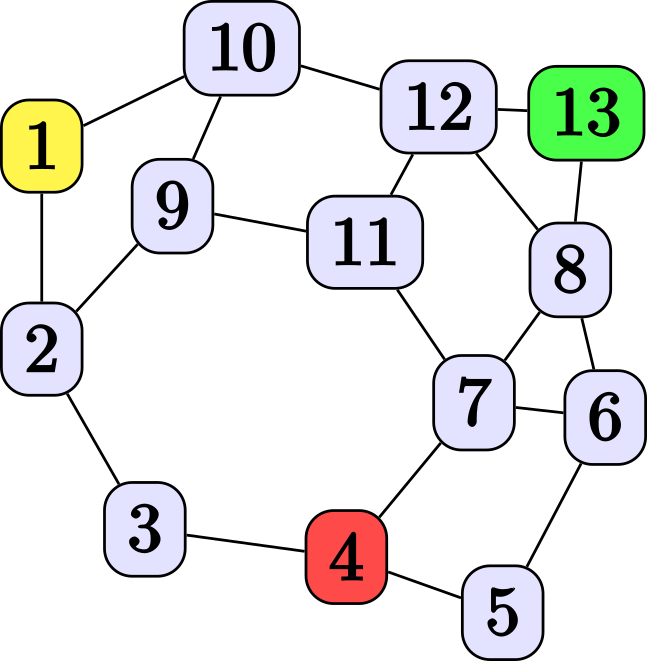}
        \vfill
        \caption{Road network N=13}
        \label{fig:roadnet}
    \end{subfigure}%
    \hfill
    \begin{subfigure}[b]{0.50\linewidth}
        \centering
        \includegraphics[width=0.6\linewidth]{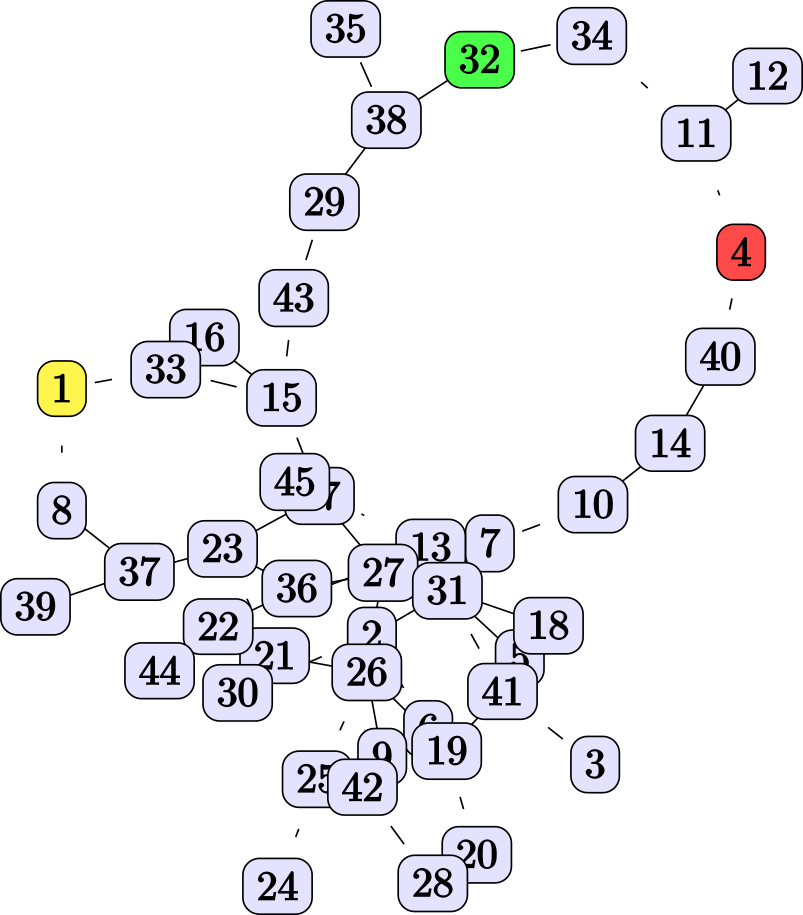}
        \caption{Road network N=45}
        \label{fig:med_roadnet}
    \end{subfigure} 
    \caption{Example road networks. UGV starts at yellow, Pursuer beings at red, and the exit is green.}
    \vspace{-0.2cm}
\end{figure}

The road network is represented as shown in Fig.~\ref{fig:roadnet}. The UGV begins at the yellow node (node 1), the pursuer begins at the red node (node 4), and the desired exit is indicated by the green node (node 13). The problem is defined by the parameters listed in Table~\ref{tab:params}

\tymin=80pt
\begin{table}[tbp]
\footnotesize
\caption{Table of parameters: simplified VIP escort problem}
\label{tab:params}
\begin{tabulary}{\linewidth}{LL}
\hline
Parameter    & Description\\
\hline
$p_{trans}$    & The transition probability of the UGV. This is the probability that the UGV will move in the desired direction when attempting to move. There is a probability of $1-t_prob$ that it will go to a different neighboring cell. \\
$d$            & MDP discount factor\\
$N$            & The number of nodes included in the road\\
% Mean Degree  & The target mean degree to which random networks are generated\\
$e_{mcts}$     & The exploration constant parameter of the MCTS.\\
$d_{mcts}$     & The depth of the MCTS tree\\
$its_{mcts}$   & The number of Monte-Carlo simulations to run to find the policy\\
$rwd_{exit}$   & The reward for the UGV successfully exiting the road network\\
$rwd_{caught}$ & The reward for the UGV being caught by the pursuer\\
$rwd_{sense}$  & The reward for making a movement\\
\hline
\end{tabulary}
\end{table}
\begin{table*}
    \footnotesize
    \centering
    \caption{Parameters used for the different experiments}
    \label{tab:exps}
    \begin{tabular}{llcccccccccc} \toprule
        &\multicolumn{10}{c}{Parameters} \\ \cmidrule(r){3-12}
        \#  & Variable(s) & Network &$p_{trans}$&$d$&$N$&$e_{mcts}$&$d_{mcts}$&$its_{mcts}$&$rwd_{exit}$&$rwd_{caught}$&$rwd_{sense}$ \\ \midrule
        1 & ${d_{mcts}}$ & Fig.~\ref{fig:roadnet} & $0.7$ & $0.90$ & 13 & $[1000.0]$ & $[1:1:10]$ & 100 & 2000 & -2000 & -200\\
        2 & ${d_{mcts}}$ & Fig.~\ref{fig:med_roadnet} & $0.7$ & $0.95$ & 45 & $[2000.0]$ & $[1:3:28]$ & 1000 & 2000 & -2000 & -200\\
        3 & ${p_{trans}}$& Fig.~\ref{fig:roadnet} & $[0.0,1.0]$ & $0.95$ & 13 & $[1000.0]$ & $[8,3,1]$ & 1000 & 2000 & -2000 & -100\\
        4 & ${p_{trans},e_{mcts}}$ & Fig.~\ref{fig:roadnet} & $[0.0,1.0]$ & $0.95$ & 13 & $[10.0,1000.0]$ & $[8,3,1]$ & 1000 & 2000 & -2000 & -100\\
    \end{tabular}
    \vspace{-0.3cm}
\end{table*}

Three separate evaluations were completed. First, \xQ{} was calculated for MCTS solvers with varying depth parameters (all others held constant). Second, \xQ{} was evaluated for a candidate solver with varying task parameters. Finally, \xQ{} was evaluated for a candidate solver with varying task \emph{and} solver parameters.

\subsection{Varying A Solver Parameter}
This evaluation involved experiments 1 and 2 from Table~\ref{tab:exps}. MCTS Solvers of varying depths were used to find solutions to each of the two networks. In each case one of the solvers was chosen as the trusted one (i.e. chose a `good' solver). In the case of experiment 1 \solvestar{} was the $d_{mcts}=9$ solver, and in experiment 2 \solvestar{} was the $d_{mcts}=25$ solver. A surrogate \surrogate{} was not used for this evaluation, instead \rwdstar{} was calculated directly and used for comparison.

\begin{figure}[tbp]
    \centering
    \begin{subfigure}[b]{0.98\linewidth}
        \centering
        \includegraphics[width=1.0\linewidth]{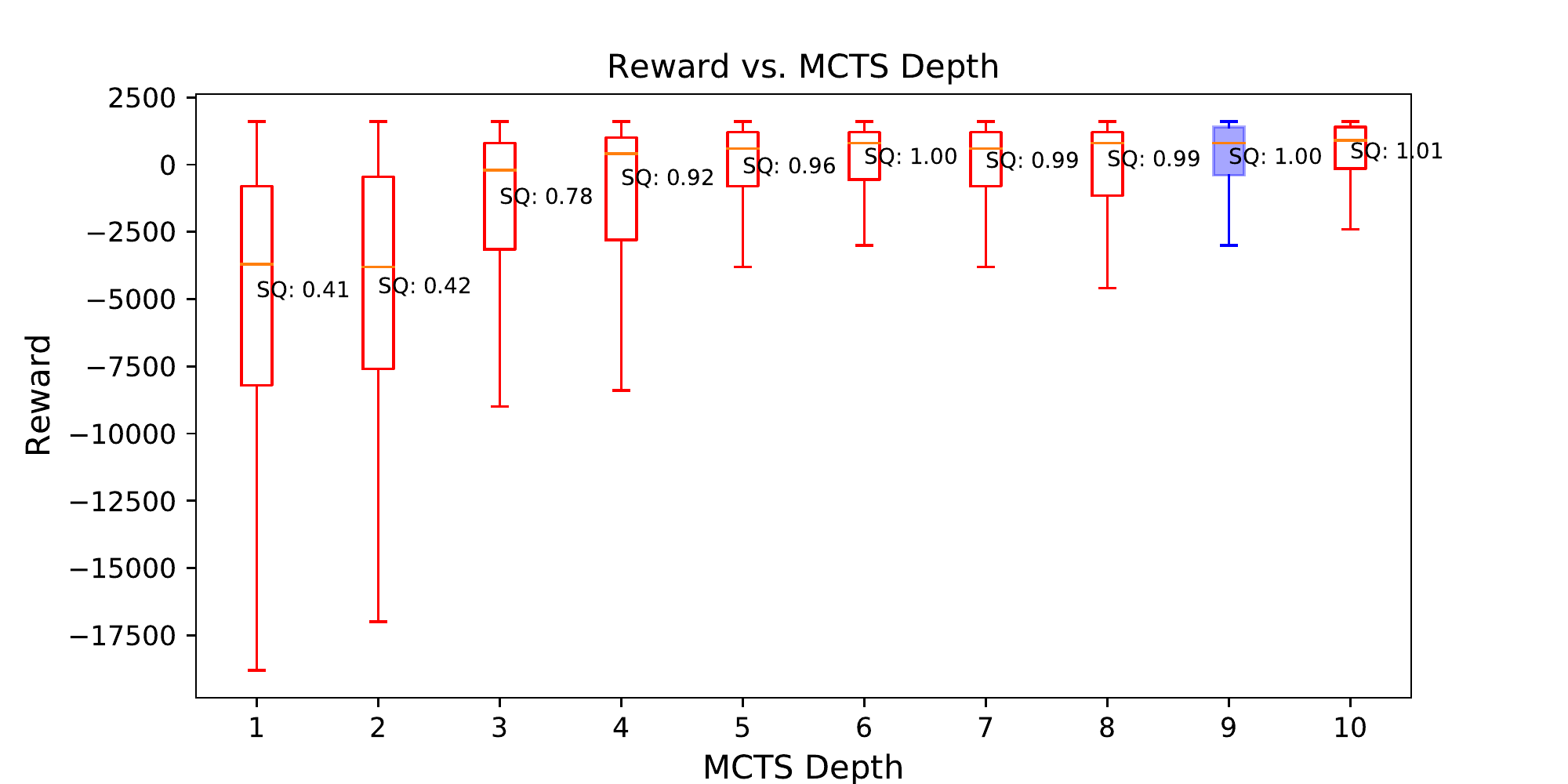}
        \vfill
        \caption{Experiment 1}
        \label{fig:mcts_d}
    \end{subfigure}%
    \hfill
    \begin{subfigure}[b]{0.98\linewidth}
        \centering
        \includegraphics[width=1.0\linewidth]{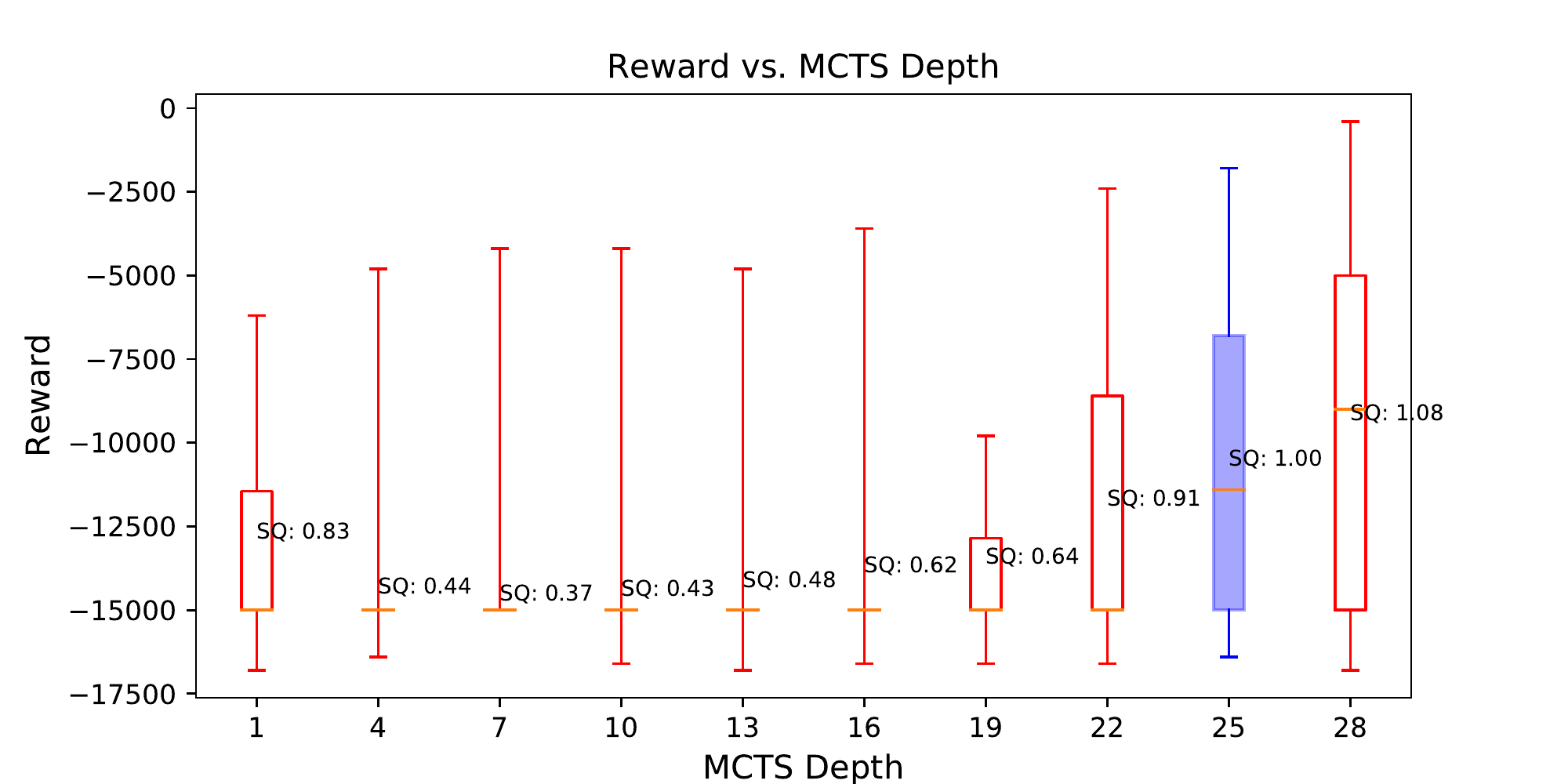}
        \caption{Experiment 2}
        \label{fig:mcts_d_med}
    \end{subfigure} 
    \caption{Experiment results. \xQ{} is calculated w.r.t \solvestar{} highlighted in blue.}
    \vspace{-0.5cm}
\end{figure}

The results for experiment 1 are shown in Fig.~\ref{fig:mcts_d}. As expected \xQ{} between \solvestar{} and itself is 1.0. We see that candidate solvers depth 6 through 10 are about equivalent to \solvestar{} (and each other), which indicates they are similarly capable of solving the problem. Whereas candidate solvers with depth 1 through 3 are much less capable than the trusted solver.

The results of experiment 2 are shown in Fig.~\ref{fig:mcts_d_med}, where \solvestar{} with $d_{mcts}=5$ has been selected. According to the plot only solvers of depth 22, and 28 have similar capability to \solvestar{}. Note, that the $d_{mcts}=1$ solver has $x_Q=0.83$, this is most possibly due to the fact that the depth 1 solver will make decisions based on little foresight, while solver with depths from 4 to 19 have enough foresight to hesitate and accumulate negative rewards from not moving quickly.

An important insight is that while the $d_{mcts}=9$ solver is very capable of solving the small network from Fig.~\ref{fig:roadnet}, that performance does not extend to the medium sized network from Fig.~\ref{fig:med_roadnet}. The $d_{mcts}=7 \text{ and } 10$ solvers are very incapable compared to the trusted solver of $d_{mcts}=25$, this is reflected by the value of \xQ.

\subsection{Varying A Task Parameter}
Experiment 3 was used for this evaluation, where \solvestar{} is a depth 8 solver, while the two candidate solvers were depth 3 and depth 1. The surrogate model \surrogate{} was learned using two generic deep neural networks with three hidden layers. One network to learn to predict the mean reward, and the other to predict the standard deviation (\surrogate{} here uses the output of two models). In a fairly simple problem such as this one a DNN of this configuration is likely overkill, but it demonstrates the possibility of using an arbitrarily complex black box model for \surrogate.

\begin{figure}[tbp]
    \centering
    \begin{subfigure}[b]{0.5\linewidth}
        \centering
        \includegraphics[width=1.1\linewidth]{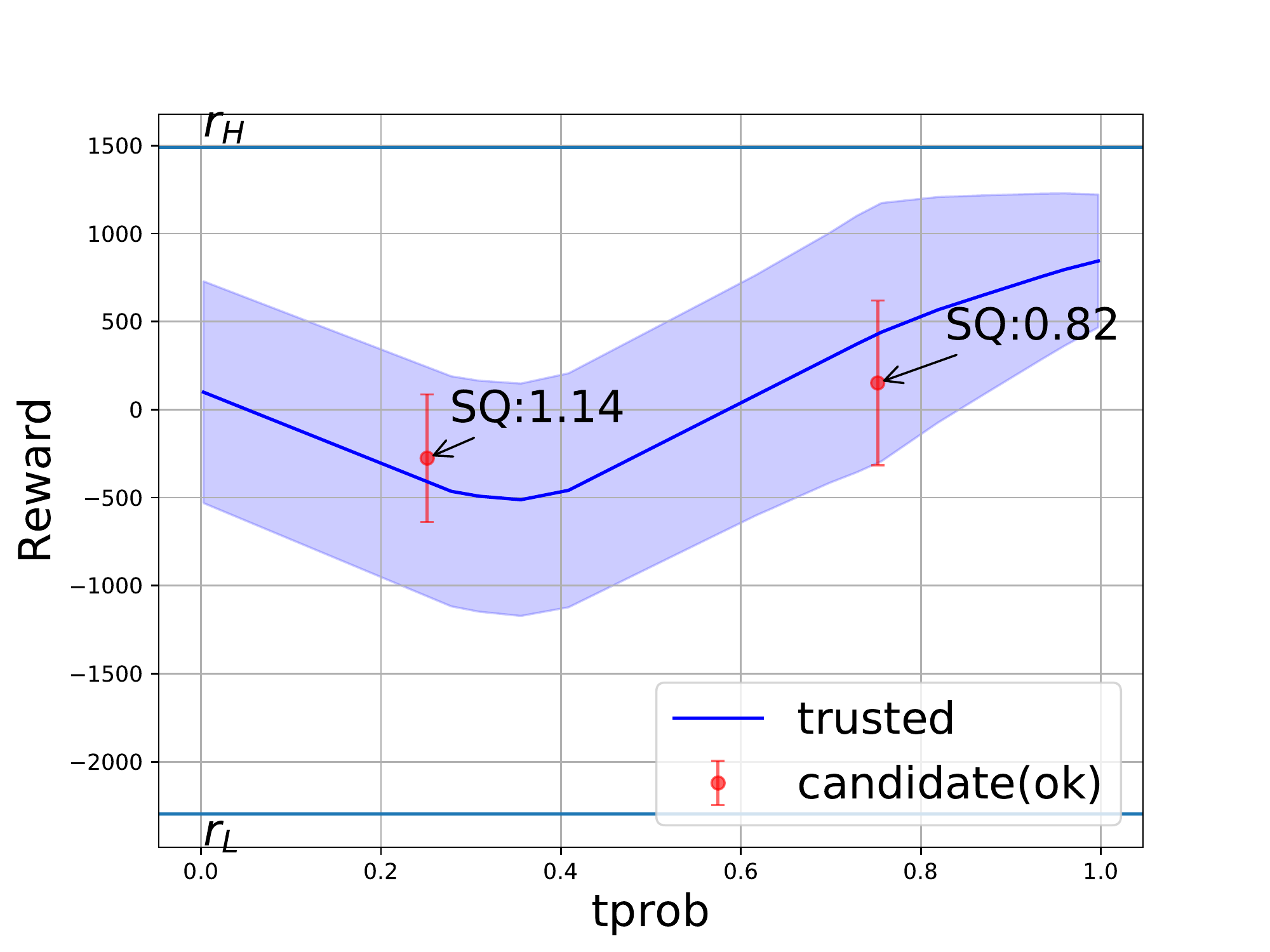}
        \vfill
        \caption{\solve{} depth 3}
        \label{fig:tprob_ok}
    \end{subfigure}%
    \hfill
    \begin{subfigure}[b]{0.5\linewidth}
        \centering
        \includegraphics[width=1.1\linewidth]{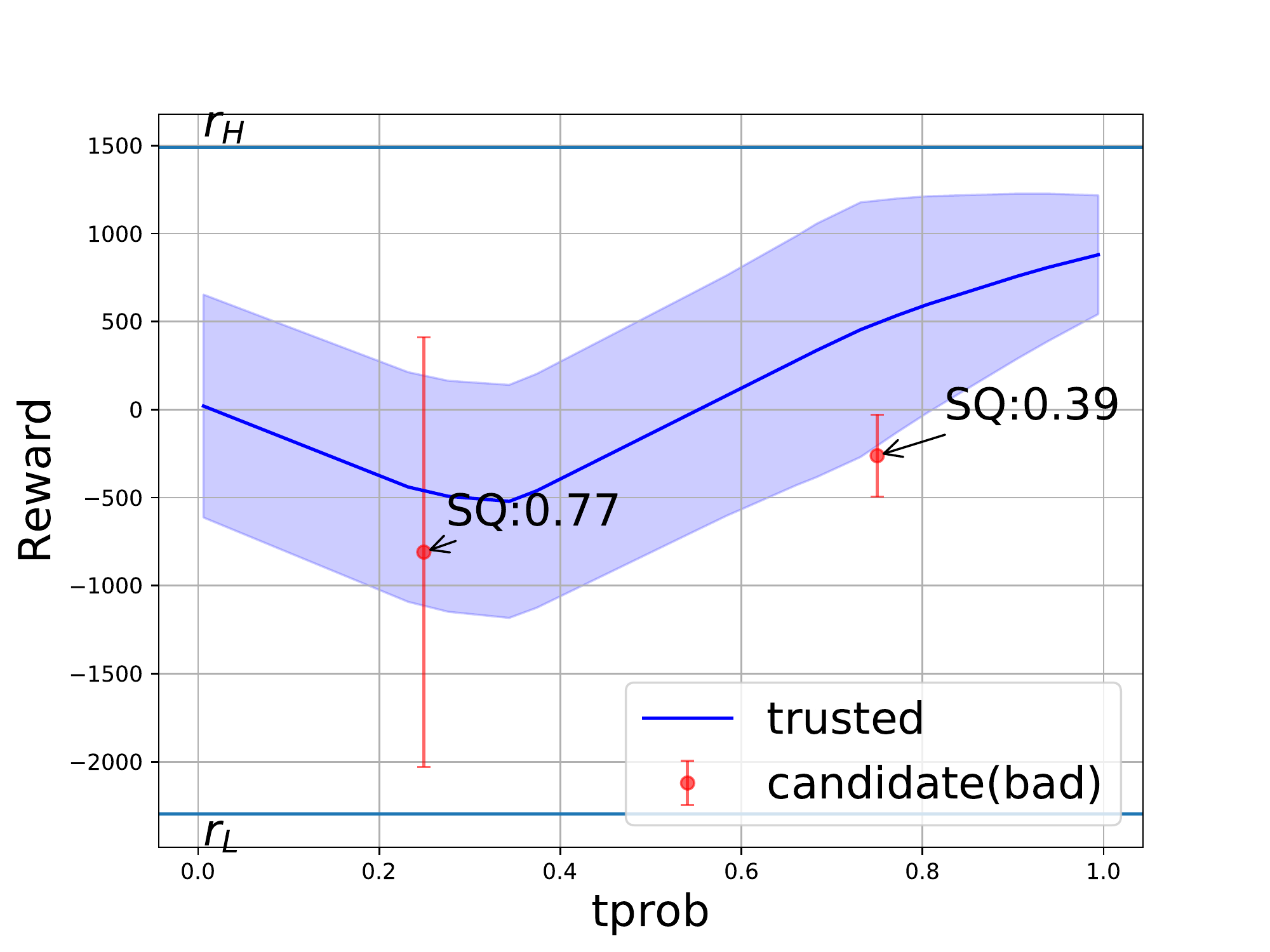}
        \caption{\solve{} depth 1}
        \label{fig:tprob_bad}
    \end{subfigure} 
    \caption{Comparison of \solve{} to \solvestar{} of depth 8.}
    \vspace{-0.5cm}
\end{figure}

Figure~\ref{fig:tprob_ok} shows the results for \solve{} with $d_{mcts}=3$ at two different values of $p_{trans}$. At $p_{trans}=0.25$ \solve{} is slightly more capable than \solvestar{}. Whereas, at $p_{trans}=0.75$ \solve{} is slightly less capable.

Figure~\ref{fig:tprob_bad} shows the results for the candidate solver with $d_{mcts}=1$ at two different values of $p_{trans}$. At $p_{trans}=0.25$ the candidate solver is moderately less capable than the trusted solver. Whereas, at $p_{trans}=0.75$ the candidate solver is slightly much less capable. These values correspond to expected behavior of \xQ.

\subsection{Varying Task and Solver Parameters}
\begin{figure}[tbp]
    \centering
    \begin{subfigure}[b]{0.98\linewidth}
        \centering
        \includegraphics[width=0.8\linewidth]{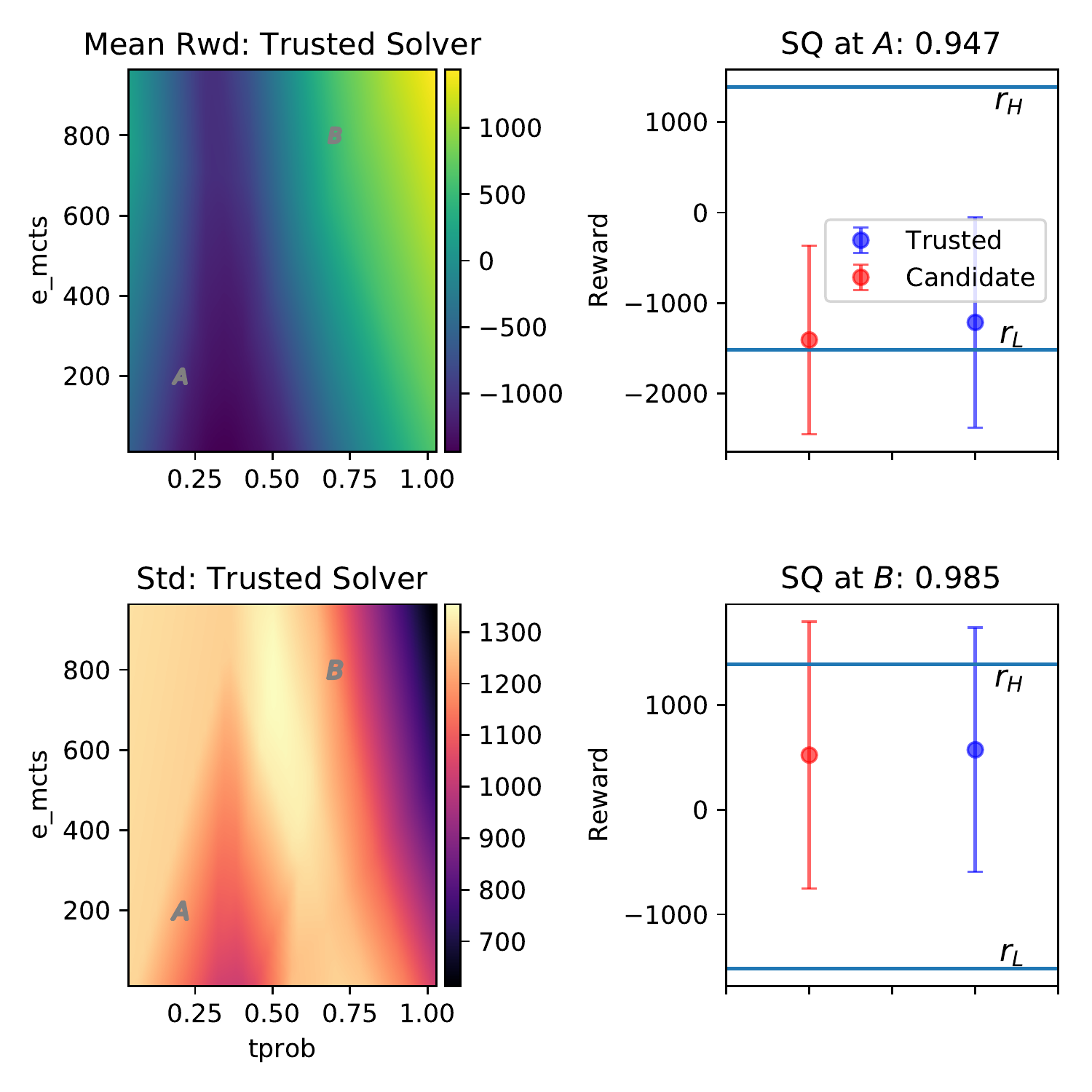}
        \vfill
        \caption{Candidate solver depth 3}
        \label{fig:tprob_emcts_ok}
    \end{subfigure}%
    \hfill
    \begin{subfigure}[b]{0.98\linewidth}
        \centering
        \includegraphics[width=0.8\linewidth]{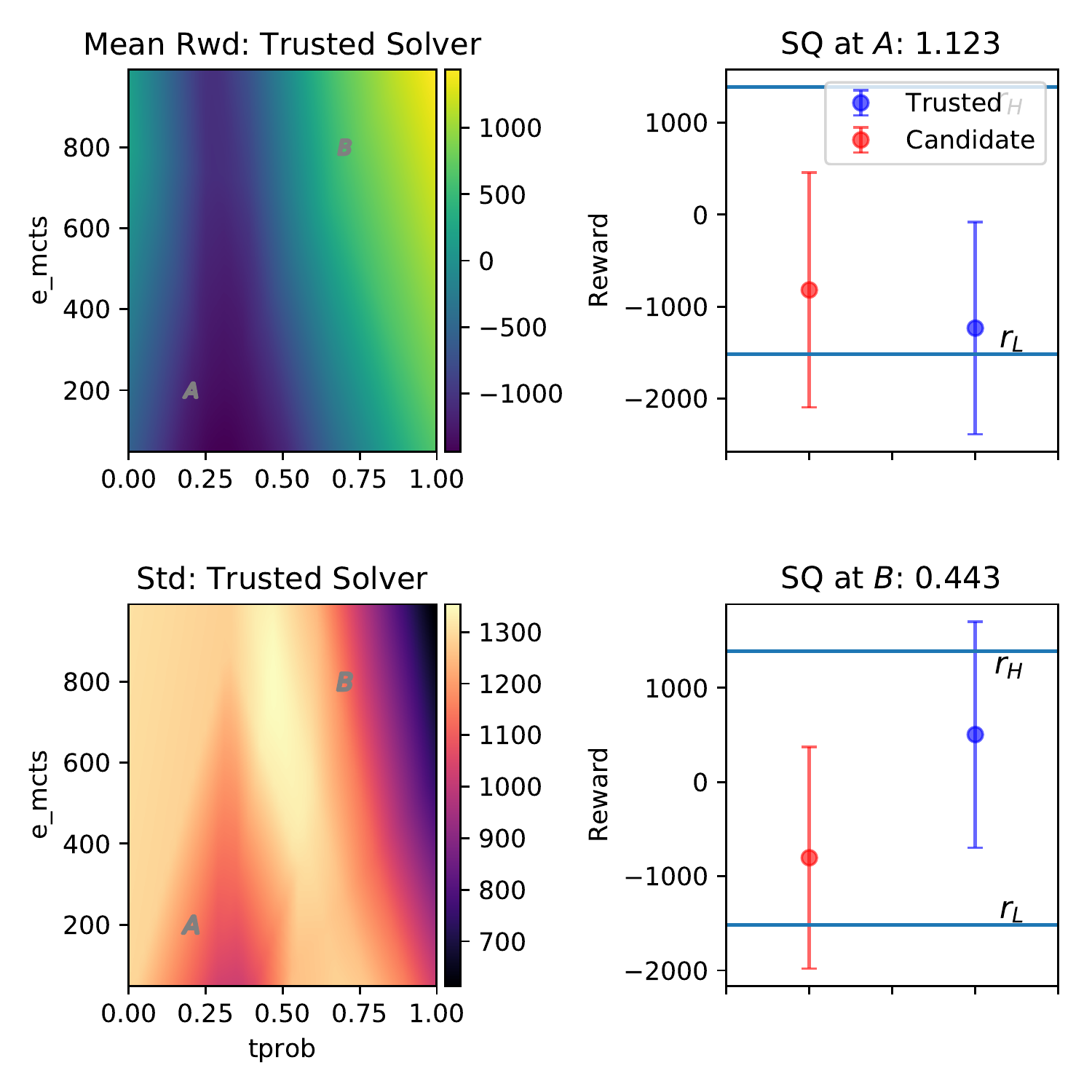}
        \caption{Candidate solver depth 1}
        \label{fig:tprob_emcts_bad}
    \end{subfigure}
    \caption{Comparison of \solve{} to \solvestar{} of depth 8. The top-left shows the mean reward of \solvestar{}, the bottom-left is the std. of \solvestar{}. Figures on RHS show \xQ{} (SQ) at points A, and B.}
    \vspace{-0.5cm}
\end{figure}

Experiment 4 was used for this evaluation, where \solvestar{} is a depth 8 solver, while the two candidate solvers were depth 3 and depth 1. Both $p_{trans}$ and $e_{mcts}$ were variable for the experiments. The results are found in Figs.~\ref{fig:tprob_emcts_ok} and \ref{fig:tprob_emcts_bad}. The surrogate \surrogate{} used here is the same as in the previous evaluation (i.e. two DNNs to predict mean and standard deviation).

Figure~\ref{fig:tprob_emcts_ok} shows the results for \solve{} with $d_{mcts}=3$ at two different points of interest. At A \solve{} is slightly less capable than \solvestar{}, \xQ{} is similar at B as well.

Figure~\ref{fig:tprob_emcts_bad} shows the results for \solve{} with $d_{mcts}=1$ at two different points of interest. At A \solve{} is slightly more capable than \solvestar{}. Whereas, at B \solve{} is moderately worse than \solvestar{}.

\section{Conclusions} \label{sec:conclusions}
Unmanned autonomous physical systems are able to tackle complex decision making problems for high-consequence applications, but in order to be able to reduce the amount of supervision required these systems need to be able to perform self-assessment, or introspection. We draw on \emph{Factorized Machine Self-Confidence (\famsec)} which is a framework of self-assessments that enable an APS to quantify its own capabilities.

Specifically, herein, we have motivated and derived a one of the factors of \famsec{} called `Solver Quality' (\xQ) that indicates the ability of some solver to perform on a given task. Calculating \xQ{} relies only on a supervised model of a trusted solver, and simulated reward distributions of candidate solvers. This approach was inspired by literature on empirical hardness models (EHMs). We have shown by numerical experiments that \xQ{}, as derived here, meets the desired criteria.

Concerning \xQ{}, it remains to be seen, and is currently left for future work, whether it actually helps users to be able to understand the capabilities and limitations of the APS. Evaluations with human participants are required.

The simulations run so far have not directly considered `different classes' of solvers, however as \xQ{} only depends on reward distributions \rwd{}, and \rwdstar{} this is not a limitation. Also, since the calculation of \xQ{} generally depends on \rwdstar{} predicted from \surrogate{} it would be prudent to enable the surrogate to predict \rwdstar{} as well as an associated uncertainty in order to have an indication of where \surrogate{} can be trusted.

Another direction for future work is to develop approaches for the remaining three \famsec{} factors. Each of the individual factors reflects a critical meta-assessment of the competency of the APS.

\bibliographystyle{ACM-Reference-Format}
\bibliography{References}
%\newpage
%\input{"appendix.tex"}

\end{document}